\pdfoutput=1

\documentclass[11pt]{article}

\usepackage[]{acl2021}

\usepackage{times}
\usepackage{latexsym}

\usepackage[T1]{fontenc}

\usepackage[utf8]{inputenc}

\usepackage{microtype}

\usepackage{makecell}

\usepackage{microtype}

\usepackage{graphicx}
\usepackage{paralist}
\usepackage{url}
\usepackage{xspace}
\usepackage{amsfonts,amsmath}
\usepackage{booktabs}
\usepackage{setspace}
\usepackage{multirow}
\usepackage{multicol}
\usepackage{subfigure}

\usepackage{hyperref}
\usepackage{amssymb}
\usepackage{pifont}
\usepackage{color, colortbl}
\usepackage{soul}
\usepackage{balance}

\newtheorem{definition}{Definition}

\newtheorem{proposition}{Proposition}

\newcommand*{\scale}[2][4]{\scalebox{#1}{$#2$}}

\makeatletter
\newcommand{\thickhline}{%
    \noalign {\ifnum 0=`}\fi \hrule height 1pt
    \futurelet \reserved@a \@xhline
}
\makeatother

\newcommand\blfootnote[1]{%
  \begingroup
  \renewcommand\thefootnote{}\footnote{#1}%
  \addtocounter{footnote}{-1}%
  \endgroup
}

\definecolor{highlight}{gray}{0.85}
\newcommand{\hlt}[1]{{\sethlcolor{highlight}\hl{#1}}}

\title{\vspace*{-0.5in}
{\small \hfill Accepted to EMNLP 2021} \\
\vspace{0.35in}
Sentence-Permuted Paragraph Generation}

\author{{\bf Wenhao Yu$^{\dag}$, Chenguang Zhu$^{\ddag}$, Tong Zhao$^{\dag}$, Zhichun Guo$^{\dag}$, Meng Jiang$^{\dag}$} \\
$\dag$University of Notre Dame $\ddag$Microsoft Cognitive Services Research \\
{\tt $\dag$\{wyu1, tzhao2, zguo5, mjiang2\}@nd.edu} \\
{\tt $\ddag$ chezhu@microsoft.com}
}

\begin{document}
\maketitle

\blfootnote{$\S$ Our code and output files are available at \url{https://github.com/wyu97/permgen}. }

\begin{abstract}
Generating paragraphs of diverse contents is important in many applications.
Existing generation models produce similar contents from homogenized contexts due to the fixed left-to-right sentence order.
Our idea is permuting the sentence orders to improve the content diversity of multi-sentence paragraph.
We propose a novel framework \textit{PermGen} whose objective is to maximize the expected log-likelihood of output paragraph distributions with respect to all possible sentence orders.
\textit{PermGen} uses hierarchical positional embedding and designs new procedures for both training phase and inference phase.
Experiments on three paragraph generation benchmarks demonstrate \textit{PermGen} generates more diverse outputs with a higher quality than existing models.

\end{abstract}

\section{Introduction}
\label{sec:introduction}


Paragraph generation is an important yet challenging task.
It requires a model to generate informative and coherent long text that consists of {multiple sentences} from free-format sources such as a topic statement or some keywords~\cite{guo2018long}.
Typical paragraph generation tasks include story generation~\cite{fan2018hierarchical}, news generation~\cite{leppanen2017data}, scientific paper generation~\cite{koncel2019text}, etc.
Recent advances in natural language generation models such as Transformer~\cite{vaswani2017attention} and BART~\cite{lewis2019bart} have demonstrated attractive performance of generating text paragraphs.


\begin{figure}[t]
    \centering
    {\includegraphics[width=0.48\textwidth, height=100pt]{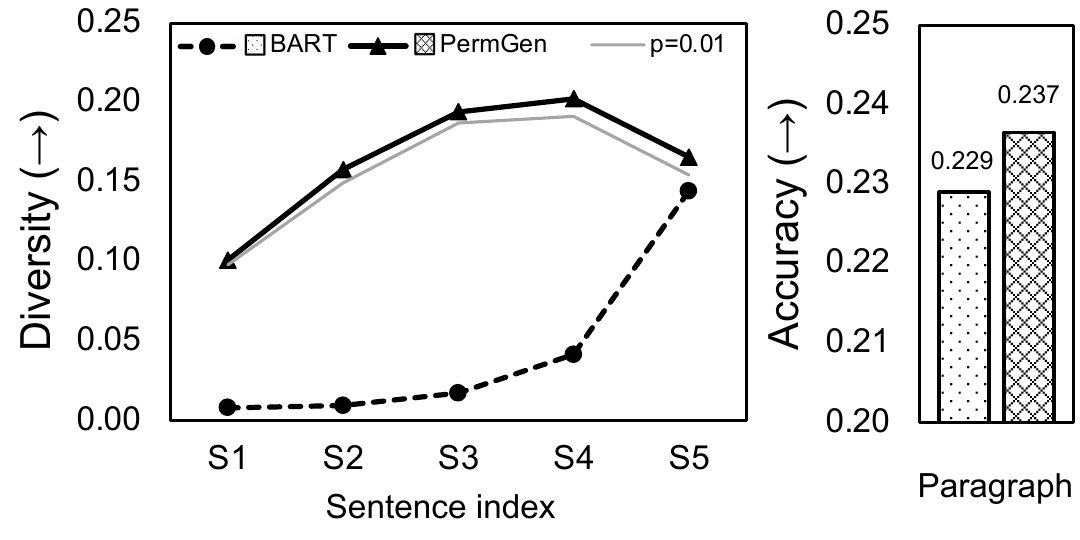}}
    \vspace{-0.3in}
    \caption{Left: Diversity of each generated story sentence at different positions (1 to 5) in ROCStories' test set, measured by averaged 1-Self-BLEU \cite{zhu2018texygen}. Our \textit{PermGen} produces contents of higher diversity at all positions, while BART (dashed line) produces diverse outputs only at the end of story. With p-value$<$0.01, \textit{PermGen} has higher diversity than the the grey line.
    Right: \textit{PermGen} outperforms BART in the accuracy of generated stories measured by BLEU-4.
    }
    \label{fig:intro}
    \vspace{-0.2in}
\end{figure}

An important desired property of model-generated paragraphs is diversity -- given the same source, an intelligent model is expected to create a variety of paragraphs in terms of content, semantic style, and word variability~\cite{li2016diversity,ippolito2019comparison}.
For example, a story generation model should narrate a plot with different storylines~\cite{clark2018creative}; a scientific paper generation model should suggest diverse contents to spark new ideas~\cite{wang2019paperrobot}.
In order to create diversity, controllable methods~\cite{zhao2017learning,cho2019mixture,yu2020survey} used additional inputs (e.g., aspects, styles). Sampling decoding algorithms~\cite{radford2019language,holtzman2020curious} searched next tokens widely from a vocabulary.
However, existing models struggled to produce multi-sentence paragraphs of diverse contents, because they relied on the homogeneity of contexts (e.g., similar story beginnings) caused by the conventional autoregressive framework with fixed left-to-right sentence order (i.e., S1$\rightarrow$S2$\rightarrow$S3).


As an example, Figure~\ref{fig:intro} evaluates the diversity of each generated sentence at different positions of the story in ROCStories~\cite{mostafazadeh2016corpus} by different models. As shown, BART (dashed line) tends to generate stories of very similar beginning and middle parts and only produce diverse text near the end of a story. 
This phenomenon stems from the fact that the left-to-right generation leads to homogeneity of context to the left, reducing the diversity of the generated paragraph.


Our idea is permuting the sentence orders in paragraph generation, while sticking with the left-to-right scheme to generate tokens in each sentence.
It has two advantages.
First, it provides an output sentence with a variety of contexts (and possibilities) from different orders. For example, creating the story ending first can probably produce a completely different story from generating the beginning first.
Second, it retains the benefit of autoregressive model that originates from the word-by-word nature of human language production.
So the coherence within sentences can be maintained, avoiding the harm of incomplete semantics from token-level permutation~\cite{shen2020blank}.

In this work, we propose a sentence-permuted paragraph generation framework called \textit{PermGen}.
Instead of using the fixed forward order, \textit{PermGen} maximizes the expected log-likelihood of the distribution in output paragraph \textit{w.r.t.} all possible sentence orders. The optimization is based on $\pi$-SGD~\citep{murphy2019janossy} which has guaranteed convergence property.
Furthermore, \textit{PermGen} employs a novel hierarchical position encoding scheme to represent the positions of tokens in permuted sentences. 
\textit{PermGen} can be initialized with any Transformer-based models 
and any decoding algorithms such as beam search and nucleus sampling~\cite{holtzman2020curious}. 

We conduct experiments on three paragraph generation tasks: story generation, news generation, and paper abstract generation. Results show that \textit{PermGen} can significantly improve the diversity of generated texts and achieve higher accuracy. Particularly, as shown in Figure \ref{fig:intro}, \textit{PermGen} model can improve diversity for sentences at all positions while also improving the accuracy.
Besides, we observe consistent improvements on both accuracy and diversity when \textit{PermGen} is coupled with various pre-trained models and decoding algorithms.

\section{Related Work}
\label{sec:related}
\noindent{\textbf{Paragraph Generation.}}
The source can be either structured or unstructured such as database records~\cite{puduppully2019data}, knowledge graphs~\cite{zhao2020graph}, images~\cite{ippolito2019comparison}, and keywords~\cite{yao2019plan}.
The expected outputs typically are stories~\cite{guan2019story,yao2019plan}, essays~\cite{yang2019enhancing}, news articles~\cite{dong2021injecting}, or scientific papers~\cite{hua2019sentence,koncel2019text}.
This task poses unique challenges as it aims at generating coherent and diverse long-form texts.
Our framework can use various forms of input such as a story title, keywords, and keyphrases, which can be generalized to broad domains.

\vspace{0.02in}
\noindent{\textbf{Diverse Text Generation.}} 
Generating diverse sequences is of crucial importance in many text generation applications that exhibit semantically \textit{one-to-many} relationships between source and the target sequences, such as machine translation~\cite{shen2019mixture,lachaux2020target}, summarization~\cite{cho2019mixture}, question generation~\cite{wang2020diversify}, and paraphrase generation~\cite{qian2019exploring}.  
Methods of improving diversity in text generation that have been widely explored from different perspectives in recent years.
Sampling-based decoding is one of the effective solutions to improve diversity~\cite{fan2018hierarchical,holtzman2020curious}, e.g., nucleus sampling~\cite{holtzman2020curious} samples next tokens from the dynamic nucleus of tokens containing the vast majority of the probability mass, instead of aiming to decode text by maximizing the likelihood.
Another line of work focuses on introducing random noise~\cite{gupta2018deep} or changing latent variable~\cite{lachaux2020target} to produce uncertainty, e.g., \citet{gupta2018deep} employ a variational auto-encoder framework to generate diverse paraphrases according to the input noise.
In addition, \citet{shen2019mixture} adopt a deep mixture of experts (MoE) to diversify machine translation, where a minimum-loss predictor is assigned to each source input; \citet{shi2018toward} employ inverse reinforcement learning for unconditional diverse text generation.

\vspace{0.02in}
\noindent{\textbf{Dynamic Order Generation.}} These methods have two categories.
First, non-autoregressive generation 
is an emerging
topic and
commonly used in machine translation~\cite{gu2018non,ren2020study}. They generate all the tokens of a sequence in parallel, resulting in faster generation speed. However, they perform poorly for long sentences due to limited target-side conditional information~\cite{guo2019non}.
Second, 
insertion-based generation
is a partially autoregressive model that maximizes the entropy over all valid insertions of tokens~\cite{stern2019insertion}. POINTER~\cite{zhang2020pointer} inherits the advantages from the insertion operation to generate text in a progressive coarse-to-fine manner. Blank language model (BLM)~\cite{shen2020blank} provides a formulation for generative modeling that accommodates insertions of various length. 

\begin{figure*}[t]
    \centering
    {\includegraphics[width=1.0\textwidth]{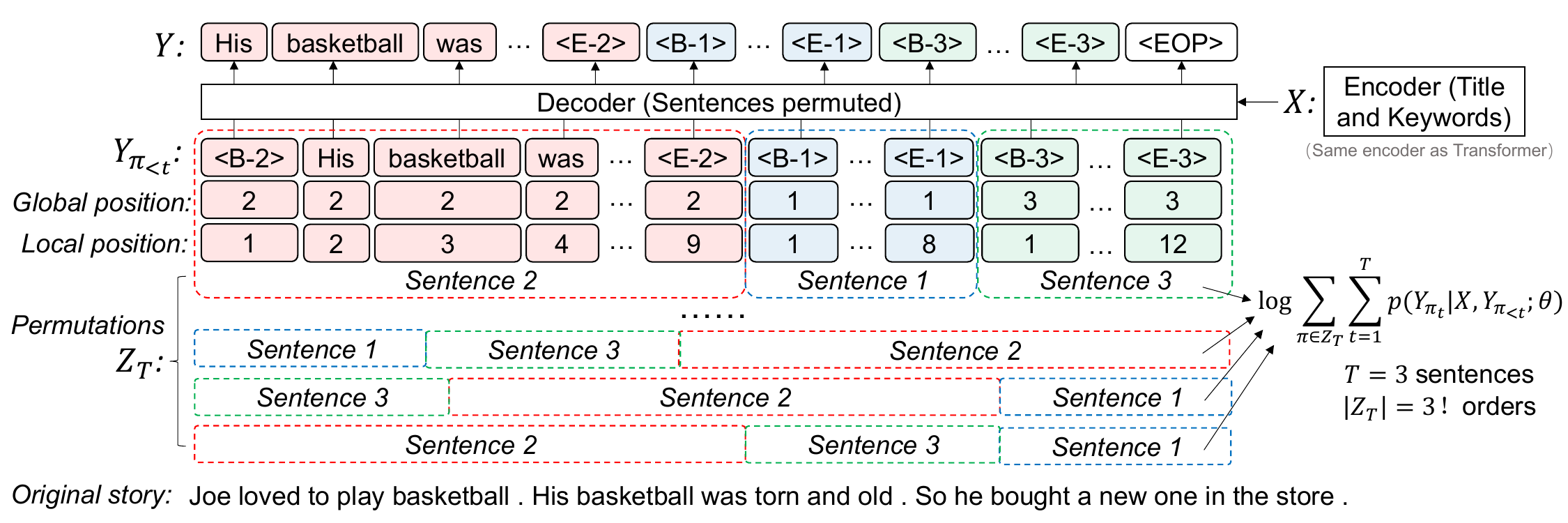}}
    \vspace{-0.35in}
    \caption{The architecture of \textit{PermGen}. The example story has 3 sentences, leading to $3!=6$ permuted sentence orders. \textit{PermGen} minimizes the overall generation loss w.r.t. all possible sentence orders.}
    \label{fig:framework}
    \vspace{-0.1in}
\end{figure*}


Different from the above methods, our \textit{PermGen} permutes the sentence orders for generating a paragraph, and it follows the left-to-right manner when producing each sentence.

\section{Preliminaries}



\noindent\textbf{Problem Definition.} Given input $X$ that can be a topic statement, some keywords, or a paper's title, the goal is to produce a paragraph $Y$ consisting of {multiple sentences} as a story, a news article, or a paper's abstract.
Suppose $Y$ has $T$ sentences, denoted by $Y= [Y_1, \cdots, Y_T]$, where $Y_t$ is the $t$-th sentence. $T$ can be easily obtained from training data
to create sentence indices. During testing, models are expected to predict the sentence indices under maximum $T$ (i.e., 10).


\subsection{Sentence-Level Transformer}

Transformer~\cite{vaswani2017attention} follows the encoder-decoder architecture~\cite{sutskever2014sequence} and uses stacked multi-head self-attention and fully connected layers for both the encoder and decoder. 
For simplicity, we represent the Transformer framework at the sentence level by using a recurrent notation that generates a probability distribution for sentence prediction by attending to both input $X$ and previous decoded sentences $Y_{<t}$. 
\begin{align}
     p(Y_t) = \mathrm{Transformer}(X, Y_{<t}).
\end{align}

\noindent where $Y_{t}$ and $Y_{<t}$ are the $t$-th sentence and sentences \textit{before} $t$-th sentence under the left-to-right manner in target output. 
Transformer eschews recurrence and instead relies on the self-attention mechanism to draw global dependencies between the input and output. During the decoding phase, Transformer can predict each token based on both the input and previously predicted tokens via attention masks to improve efficiency. 
The objective of Transformer is to maximize the likelihood under the forward autoregressive factorization:
\begin{align}
    p(Y|X; \theta) = \prod^{T}_{t=1} p(Y_t|Y_{< t}, X; \theta) .
\label{eq:Seq2Seq-loss}
\end{align}



\section{Proposed Method: \textit{PermGen}}


In a left-to-right generation scheme such as the canonical Seq2Seq design, each generated token is conditioned on left-side tokens only~\cite{sutskever2014sequence}. It ignores contextual dependencies from the right side. It also leads to limited diversity of generated text (as shown in Figure \ref{fig:intro}).
To solve this problem, our \textit{PermGen}, a novel sentence-permuted paragraph generation model, produces sentences not confined to the left-to-right order. Instead, \textit{PermGen} attempts different sentence orders and selects the best-ranked output candidate. 

As shown in Figure \ref{fig:framework}, \textit{PermGen} uses the Transformer encoder but changes the sentence orders during the decoding phase. It should be noted that \textit{PermGen} follows the left-to-right manner when generating tokens in each sentence. Thus, we represent the Transformer decoder as:
\begin{align}
    {Y_{\pi_t} = \mathrm{Transformer}(X, Y_{\pi_{<t}}, \pi),}
\label{eq:TransEncoder_2}
\end{align}
where $Y_{\pi_t}$ and $Y_{\pi_{<t}}$ are the $t$-th sentence and the sentences \textit{before} the $t$-th sentence under the permutation order $\pi$ in the target output. Taking the first permuted order in Figure \ref{fig:framework} as an example, we have $\pi=[2, 1, 3]$, $\pi_1=2$,  $\pi_3=3$, $\pi_{<3}=[2, 1]$.

We note that as \textit{PermGen} is based on the encoder-decoder Transformer architecture, which can be initialized either randomly or from a pre-trained Transformer model with the same structure. Therefore, in the experiments, we evaluate \textit{PermGen} which is i) trained from scratch, and ii) initialized with BART \citep{lewis2019bart}.
Next, we will introduce three modules of \textit{PermGen}: (1) hierarchical positional embedding, (2) sentence-permuted learning, and (3) sentence-based decoding.


  
    

\begin{figure*}[ht]
    \centering
    {\includegraphics[width=1.0\textwidth]{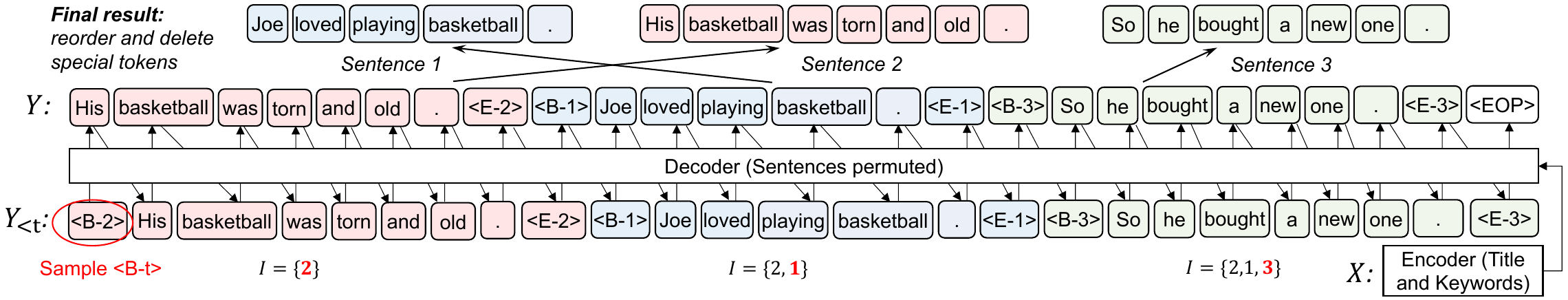}}
    \vspace{-0.25in}
    \caption{The decoding process during inference as described in Section \ref{sec:decode}. Note that the first special token (e.g., <B-2>) is sampled from \{<B-$t$>\}$_{t=1}^{T}$.
    For simplicity, positional embedding is omitted in the figure. }
    \label{fig:inference}
    \vspace{-0.1in}
\end{figure*}

\subsection{Hierarchical Positional Embedding}

In Transformer, positional embeddings are added to every token's embedding. Traditionally, the positional embedding encodes the absolute position from 1 to the sequence length to model how a token at one position attends to tokens at other positions~\cite{vaswani2017attention,lewis2019bart}.

We propose the hierarchical positional embedding 
that consists of a global position and a local position. Given a token, the global position is the position (index) of the sentence that contains this token; the local position is the position of the token in the sentence (see the two lines of position numbers in Figure \ref{fig:framework}). Given a paragraph $Y$, its embedding matrix is given below, where rows are its tokens and columns are embedding dimensions:
\vspace{-0.2in}
\begin{align}
    \textbf{Y} = \textbf{Y}_{\text{token}} + \textbf{Y}_{\text{global\_position}} + \textbf{Y}_{\text{local\_position}},
\end{align}
where $\textbf{Y}_{\text{token}}$ is the token embedding, $\textbf{Y}_{\text{global\_position}}$ and $\textbf{Y}_{\text{local\_position}}$ are the global positional embeddings and local positional embeddings.

Compared to the absolute positional embedding, the hierarchical positional embedding has two advantages.
First, the embedding of two-level positions is more informative about the paragraph structure than that of the absolute position.
Second, when we permute the sentence orders in paragraph generation, the absolute positions of tokens might not be available. 
For example, if the second sentence is generated earlier than the first sentence, the absolute positions of its tokens cannot be determined because the length of the first sentence is unknown. In comparison, hierarchical position does not have this issue.

In addition, for the $t$-th sentence in $Y$, we add two special tokens (i.e., <B-$t$> and <E-$t$>) to indicate the beginning and end of the sentence.
Thus, the decoder can determine the sentence index based on the predicted special tokens.
We also append a special token <EOP> to the paragraph to indicate the end of the generation process.

\subsection{Sentence-permuted Learning}

This module learns by varying sentence orders in paragraph generation and acts as the key component in \textit{PermGen}. For example, given a sentence order $\pi = [2, 4, 1, 5, 3]$, \textit{PermGen} first generates the second sentence from the leftmost token to the rightmost, then generates the fourth sentence, and so on.
Formally, we denote $Z_T$ as the set of all possible sentence orders, i.e., the permutations of sentence indices of length $T$. It follows that $|Z_T| = T!$.
Given input $X$ and target output paragraph $Y$ of $T$ sentences, \textit{PermGen} maximizes the following likelihood:
\begin{align}
    p(Y|X;\theta) &= \sum_{\pi \in Z_T} p (Y|X, \pi; \theta) \nonumber \\
    & = \sum_{\pi \in Z_T}\prod_{t=1}^T p (Y_{\pi_t}|X, Y_{\pi_{<t}}; \theta)
    \label{eq:original-loss} 
\end{align}
However, computing the negative log-likelihood in Eq. (\ref{eq:original-loss}) is prohibitive because
the back-propagation computational graph branches out for every permutation in the sum. 
Therefore, we apply the Jensen's inequality to lower-bound the log-likelihood:
\begin{align}
     & \scale[0.98]{\log p(Y|X;\theta)} \scale[0.98]{=}~ \scale[0.98]{\log \sum\limits_{\pi \in Z_T} \prod\limits_{t=1}^{T} p (Y_{\pi_t}|X, Y_{\pi_{<t}}; \theta)} \nonumber \\
     \scale[0.98]{\geq}~& \scale[0.98]{\log (|Z_T|) + \frac{1}{|Z_T|} \sum\limits_{\pi \in Z_T} \sum\limits_{t=1}^{T} \log p (Y_{\pi_t}|X, Y_{\pi_{<t}}; \theta)}\label{eq:new-loss} \nonumber
\end{align}
By maximizing the lower bound, we do not favor any particular sentence order, but encourage the model to generate $Y$ equally well in all orders.

Note that maximizing this lower bound is equivalent to minimizing the following expectation:
\begin{equation}
     \mathcal{J}(\theta)=  \mathbb{E}_{\pi^\prime} \Big{[}-\sum_{t=1}^{T} \log p (Y_{\pi^\prime_t}|X, Y_{\pi^\prime_{<t}}; \theta) \Big{]}. \label{eq:upper-b} 
\end{equation}

Since computing this expectation is still intractable, we apply the $\pi$-SGD~\cite{murphy2019janossy} stochastic optimization, which randomly samples a permutation for gradient computation.

\begin{definition}
    ($\pi$-SGD): Let $\mathcal{B} = \{(X^{(1)}, Y^{(1)}), $ $ \cdots ,(X^{(B)}, Y^{(B)})\}$ be a mini-batch i.i.d. sampled uniformly from the training data $D$. At step $t$, consider the stochastic gradient descent update 
    \begin{equation}
        \theta_t = \theta_{t-1} - \eta_t G_t,   
        \label{eq:pi-sgd}
    \end{equation}
    where $ G_t= - \frac{1}{B} \sum_{i=1}^{B} \nabla_\theta \sum_{t=1}^{T} \log p (Y^{(i)}| X^{(i)}, $ $\pi^\prime; \theta)$ is the gradient, and random permutations $\{\pi^\prime_i\}_{i=1}^B$ are sampled independently: $\pi^\prime_i \sim \mathrm{Uniform}(Z_T^{(i)})$. Besides, the learning rate is $\eta_t \in (0, 1)$ s.t. $\rm{lim}_{t \rightarrow \infty} \eta_t = 0 $, and $\sum_{t=1}^{\infty} \eta_t^{2} < \infty$.
    \label{def:pi-sgd}
\end{definition}

\vspace{0.02in}
We note that $\pi$-SGD is a Robbins-Monro stochastic approximation of gradient descent~\cite{robbins1951stochastic}. When it's applied to permutation sampling, the optimization almost surely converges to the optimal $\theta$, as implied by the following proposition.



\begin{proposition}
    ($\pi$-SGD Convergence): The optimization of $\pi$-SGD converges to the optimal $\theta$ for $\mathcal{J}(\theta)$ in Eq. (\ref{eq:upper-b}) with probability one.
    \label{proposition:convergence}
\end{proposition}
Proof: We refer to Prop.2.2 in \citet{murphy2019janossy}.

\subsection{Sentence-based Decoding}


\label{sec:decode}
\noindent In decoding, \textit{PermGen} adopts the following steps:
\vspace{0.02in}
\begin{compactitem}
  \item \textit{Step 1:} Initialize a set of indices of sentences that have been generated: $I = \{\}$;
  \item \textit{Step 2:} If $I = \{\}$, sample a token from \{<B-$t$> $\mid$ $t$ $\in$ \{1, $\dots$, $T$\}\}\footnote{When trying to generate multiple candidates, we use the sampling without replacement strategy. For example, if we need to generate 3 candidates each with 5 sentences, their beginning tokens can be B-1, B-3 and B-4, respectively. }; otherwise, predict a token from \{<B-$t$> $\mid$ $t$ $\in$ \{1, $\dots$, $T$\}$\setminus I$\} $\cup$ \{<EOP>\}. If the token is <EOP>, end; otherwise, append <B-$t$> to the generated text;
  \item \textit{Step 3:} Generate tokens from $\mathcal{V}\cup\{$<E-$t$>$\}$ for the $t$-th sentence in an autoregressive way, where $\mathcal{V}$ is the set of normal text tokens. Stop when <E-$t$> is generated;
  \item \textit{Step 4:} $I \leftarrow I\cup\{t\}$, then go back to Step 2.
\end{compactitem}

\vspace{0.02in}

As stated in \textit{step 2}, when <EOP> is generated, the whole generation ends. Then, the sentences in the generated paragraph can be reordered according to sentence indices $I$ and special tokens.
Note that in \textit{step 3}, since \textit{PermGen} adopts autoregressive generation, it can employ any decoding strategy such as beam search or sampling algorithm (e.g. truncated sampling~\cite{fan2018hierarchical}, nucleus sampling~\cite{holtzman2020curious}). For example, truncated sampling samples the next word from the top $k$ probable choices, instead of aiming to decode text by maximizing the likelihood.

\vspace{0.02in}
\noindent\textbf{Rank with log-probability.}
We compute the log-likelihood of each candidate as the same as in beam search~\cite{vijayakumar2016diverse} and sampling methods~\cite{holtzman2020curious}:
\begin{align}
    S_\text{prob}(Y) =
    \frac{1}{L} \sum_{l=1}^{L} \log p(y_{l}|y_1, \cdots, y_{l-1}) 
    \label{eq:beam-prob}
\end{align}
where $L$ is the total number of tokens in $Y$ and $y_l$ is the $l$-th token in generated paragraph $Y$.

\vspace{0.02in}
\noindent\textbf{Complexity reduction.}
Since the number of possible sentence orders grows as $n!$ for a $n$-sentence paragraph, exact inference is an extremely time consuming process. To reduce the complexity during inference, we 
employ an approximate inference by taking advantage of the special token prediction mentioned in \textit{step 2}. 
The special token prediction happens when a end-of-sentence (i.e., <E-$t$>) is generated. Instead of traversing each remaining possible sentence index, the model only chooses the most likely sentence index through special token predictions. 
It should be noted that we reuse the classifier in decoder by simply masking tokens \textit{not in} \{<B-$t$>\}$_{t=1}^T$, without training any new classifiers.
Therefore, the decoding time is roughly linear in the number of candidates to be generated. 



\section{Experiments}
\label{sec:Experiments}
We conduct experiments on three text generation tasks: story generation, news generation, and paper abstract generation. 
For all tasks, we compare \textit{PermGen} with multiple baseline models on diversity and accuracy of their generated texts. We also perform human evaluation on story generation.

\subsection{Tasks and Benchmarks}

\paragraph{Task 1: Story generation} In this task, models learn to generate story paragraphs from the title and multiple keywords.
We use ROCStories dataset~\cite{mostafazadeh2016corpus} and follow the same data preparation as in~\citet{yao2019plan}.
ROCStories has 98,162 / 9,817 / 9,803 paragraphs for training / development / test sets, respectively. The stories in the corpus 
capture causal and temporal commonsense relations between daily events.

\begin{table}[t]
\caption{Statistics of three datasets. ``in/out'' stands for input/output and ``sents'' stands for sentences.}
\vspace{-0.15in}
\begin{center}
\setlength{\tabcolsep}{2mm}{
\scalebox{0.85}{\begin{tabular}{l|c|c|c}
\toprule
Dataset & ROCStories & AGENDA & DailyMail \\
\midrule
\# Train & 98,162 & 38,720 & 49,102 \\
\# Dev. & 9,817 & 1,000 & 2,000 \\
\# Test & 9,803 & 1,000 & 2,000 \\
Title in input & $\surd$ & $\surd$ & $\times$ \\
Avg.in.words & 9.65 & 16.09 & 7.91 \\
Avg.out.words & 50.16 & 76.12 & 95.62 \\
Avg.out.sents & 4.92 & 3.08 & 3.88 \\
\bottomrule
\multicolumn{4}{l}{* The DailyMail dataset does not have news title for each} \\
\multicolumn{4}{l}{article. We only generate the first paragraph of news in} \\
\multicolumn{4}{l}{DailyMail. The average length is 3.88 sentences.}
\end{tabular}}}
\vspace{-0.15in}
\label{tab:datasets}
\end{center}
\end{table}

\begin{table*}[htb]
\begin{center}
\caption{Diversity (``Dist-2'': Distinct-2($\Uparrow$), ``Self-B-4'': Self-BLEU-4($\Downarrow$)) and accuracy (``B-4'': BLEU-4($\Uparrow$)) for \textit{PermGen} and baseline methods. Diversity evaluation is calculated by top-k generated candidates from \textit{beam search}.
We use bold and underline to indicate the best and best baseline performance.
}
\vspace{-0.1in}
\setlength{\tabcolsep}{1mm}{
\scalebox{0.83}{\begin{tabular}{lc||cc|c|cc|c|cc|c}
\toprule
{\multirow{3}*{Methods}} &
{\multirow{3}*{\makecell[c]{Pre-\\Train}}}  & \multicolumn{3}{c|}{{ROCStories}} & \multicolumn{3}{c|}{{AGENDA}} & \multicolumn{3}{c}{{DailyMail}} \\
& & \multicolumn{2}{c|}{Diversity} & \multicolumn{1}{c|}{Accuracy} & \multicolumn{2}{c|}{Diversity} & \multicolumn{1}{c|}{Accuracy} & \multicolumn{2}{c|}{Diversity} & \multicolumn{1}{c}{Accuracy} \\
\cmidrule{3-11}
& & Dist-2($\Uparrow$) & Self-B-4($\Downarrow$) & B-4($\Uparrow$) & Dist-2($\Uparrow$) & Self-B-4($\Downarrow$) & B-4($\Uparrow$) & Dist-2($\Uparrow$) & Self-B-4($\Downarrow$) & B-4($\Uparrow$) \\
\midrule
POINTER & $\surd$ & 0.0743 & 0.9405 & 0.0492 & 0.1898 & 0.9267 & 0.0379 & 0.1228 & 0.9619 & 0.0243 \\
BLM & $\surd$ & 0.0560 & 0.9573 & 0.1477 & 0.1465 & 0.9396 & 0.1679 & 0.0831 & 0.9889 & 0.1164 \\
GPT-2 & $\surd$ & \underline{0.0915} & \underline{0.9194} & 0.0726 & 0.1665 & 0.9331 & 0.1247 &  \underline{0.1577} & \underline{0.9287} & 0.1072 \\
BERTGen & $\surd$ & 0.0672 & 0.9456 & 0.1576 & 0.1463 & 0.9356 & 0.1462 & 0.1167 & 0.9774 & 0.1728 \\
T5 & $\surd$ & 0.0684 & 0.9403 & 0.1895 & 0.1323 & 0.9421 & 0.1688 & 0.1086 & 0.9779 & 0.1529 \\
Transformer & $\times$ & 0.0806 & 0.9341 & 0.1809 & 0.1489 & \underline{0.9265} & 0.1540 & 0.1109 & 0.9678 & 0.1496 \\
BART & $\surd$ & 0.0839 & 0.9330 & \underline{0.2445} & \underline{0.1697} & 0.9278 & \underline{0.1922} & 0.1306 & 0.9720 & \underline{0.1935} \\
\midrule
{\multirow{2}*{\textit{PermGen}}} & $\times$ & 0.0992 & 0.8548 & 0.1848 & 0.2203 & \textbf{0.5679} & 0.1678 & 0.1934 & 0.7757 & 0.1592\\
& $\surd$ & \textbf{0.1059} & \textbf{0.7993} & \textbf{0.2482} & \textbf{0.2492} & 0.5940 & \textbf{0.2059} & \textbf{0.2065} & \textbf{0.6627} & \textbf{0.1991} \\
\bottomrule
\end{tabular}}}
\label{tab:baseline}
\end{center}
\end{table*}

\begin{table*}[htb]
\begin{center}
\caption{Ablation study. Hi-BART represents BART with hierarchical positional embeddings.}
\vspace{-0.1in}
\setlength{\tabcolsep}{1mm}{
\scalebox{0.83}{\begin{tabular}{lc||cc|c|cc|c|cc|c}
\toprule
{\multirow{3}*{Methods}} &
{\multirow{3}*{\makecell[c]{Pre-\\Train}}}  & \multicolumn{3}{c|}{{ROCStories}} & \multicolumn{3}{c|}{{AGENDA}} & \multicolumn{3}{c}{{DailyMail}} \\
& & \multicolumn{2}{c|}{Diversity} & \multicolumn{1}{c|}{Accuracy} & \multicolumn{2}{c|}{Diversity} & \multicolumn{1}{c|}{Accuracy} & \multicolumn{2}{c|}{Diversity} & \multicolumn{1}{c}{Accuracy} \\
\cmidrule{3-11}
& & Dist-2($\Uparrow$) & Self-B-4($\Downarrow$) & B-4($\Uparrow$) & Dist-2($\Uparrow$) & Self-B-4($\Downarrow$) & B-4($\Uparrow$) & Dist-2($\Uparrow$) & Self-B-4($\Downarrow$) & B-4($\Uparrow$) \\
\midrule
BART & $\surd$ & 0.0839 & 0.9330 & 0.2445 & 0.1697 & 0.9278 & 0.1922 & 0.1306 & 0.9720 & 0.1935 \\
Hi-BART & $\surd$ & 0.0812 & 0.9356 & 0.2349 & 0.1673 & 0.9265 & 0.1880 & 0.1289 & 0.9705 & 0.1899 \\
\textit{PermGen} & $\surd$ & \textbf{0.1059} & \textbf{0.7993} & \textbf{0.2482} & \textbf{0.2492} & \textbf{0.5940} & \textbf{0.2059} & \textbf{0.2065} & \textbf{0.6627} & \textbf{0.1991} \\
\bottomrule
\end{tabular}}}
\label{tab:ablation}
\end{center}
\end{table*}

\paragraph{Task 2: Paper abstract generation} In this task, models need to generate paper abstracts from paper title and a list of keywords. We use the AGENDA dataset~\cite{koncel2019text} that consists of 40,720 paper titles and abstracts in the Semantic Scholar Corpus taken from the proceedings of 12 AI conferences. Each abstract is paired with several keywords. We follow the settings in~\citet{koncel2019text} to directly generate paper abstracts from the keywords. We follow the same data partition, which has 38,720 / 1,000 / 1,000 for training / development / test sets, respectively.

\paragraph{Task 3: News generation} In this task, models are trained to generate news articles from a list of keyphrases. We use DailyMail dataset~\cite{see2017get}, a corpus of online news articles. We randomly sample 53,102 news articles and extract keyphrases from each sentence using RAKE~\cite{rose2010automatic}. It contains 49,102 / 2,000 / 2,000 news articles for training / development / test sets.

\begin{figure*}[t]
    \centering
    {\includegraphics[width=1.0\textwidth]{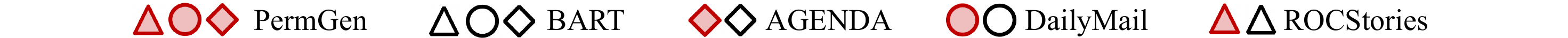}} \\
    \vspace{0.08in}
    {\includegraphics[width=1.0\textwidth]{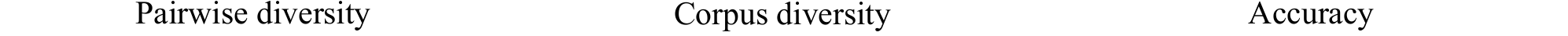}} \\
    \vspace{-0.1in}
    \subfigure[Self-BLEU-3($\Downarrow$)]
    {\includegraphics[width=0.328\textwidth]{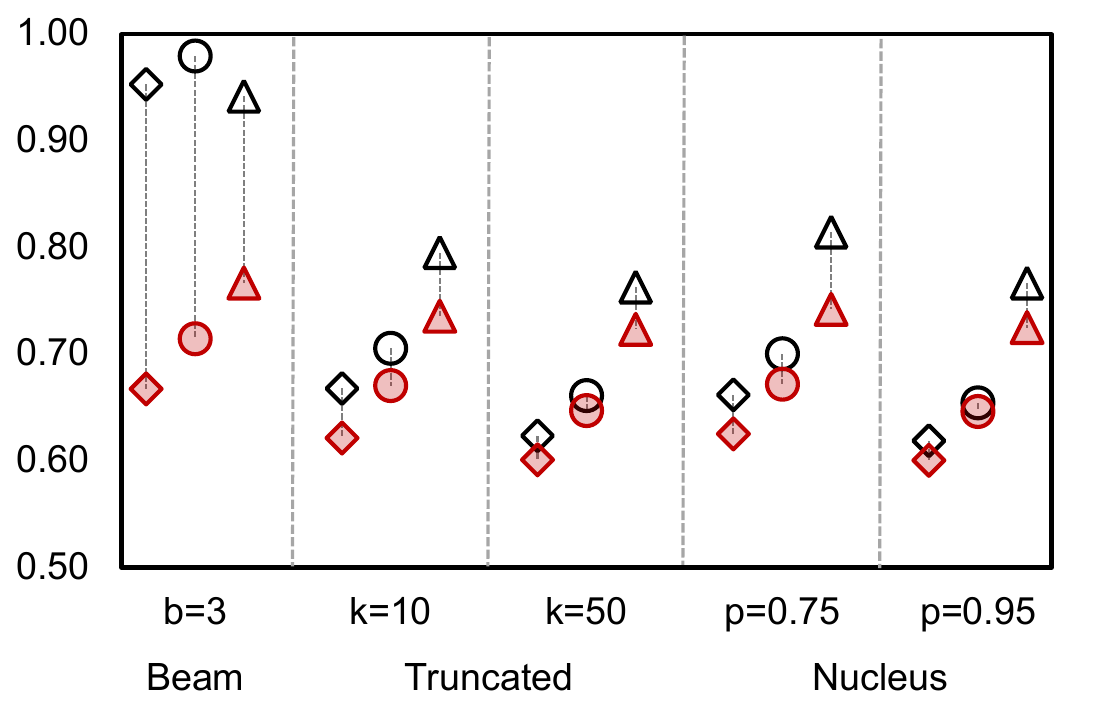}}
    \vline
    \subfigure[Distinct-2($\Uparrow$)]
    {\includegraphics[width=0.328\textwidth]{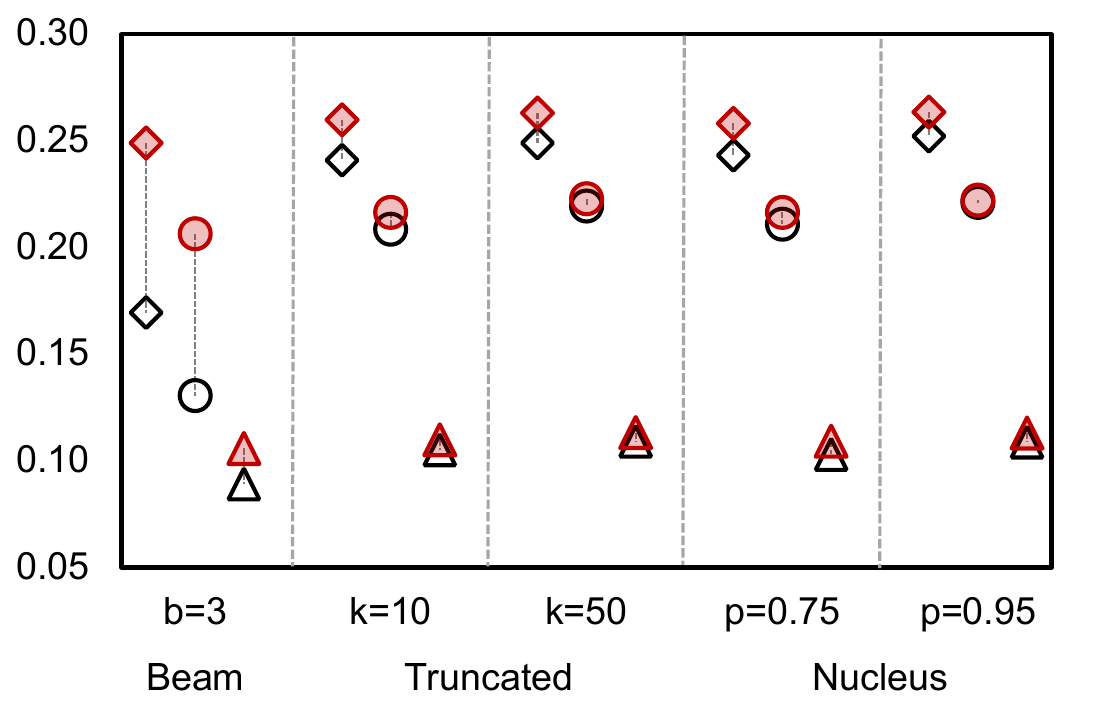}}
    \vline
    \subfigure[Top1-BLEU-4($\Uparrow$)]
    {\includegraphics[width=0.328\textwidth]{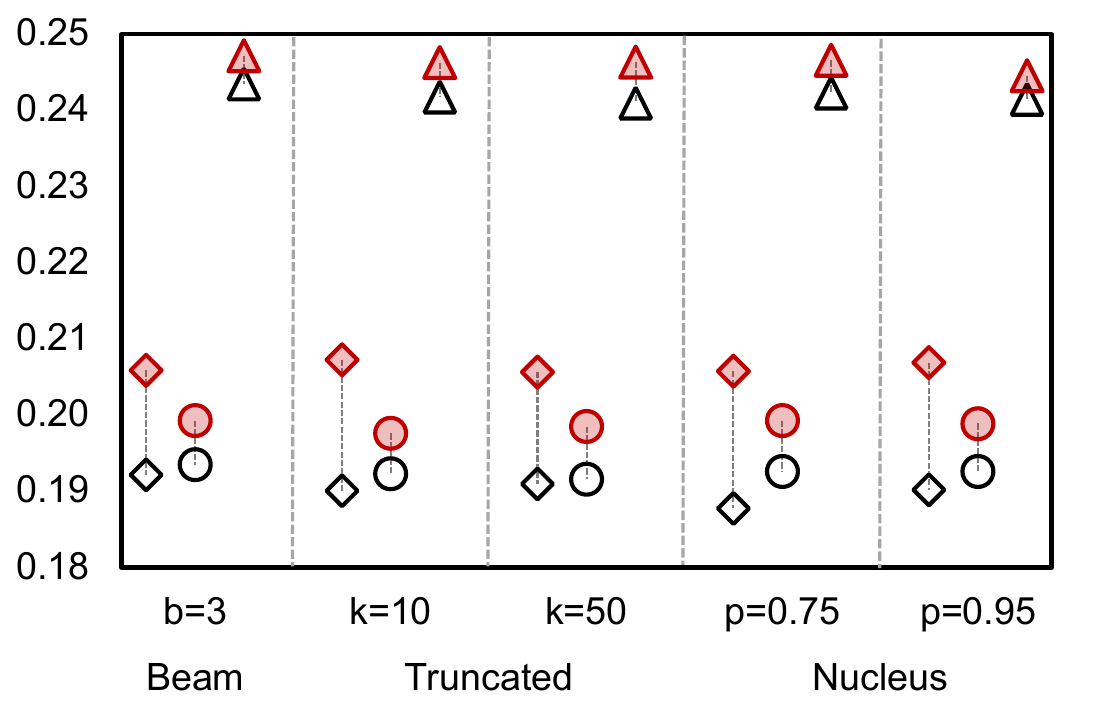}} \\
    \subfigure[Self-BLEU-4($\Downarrow$)]
    {\includegraphics[width=0.328\textwidth]{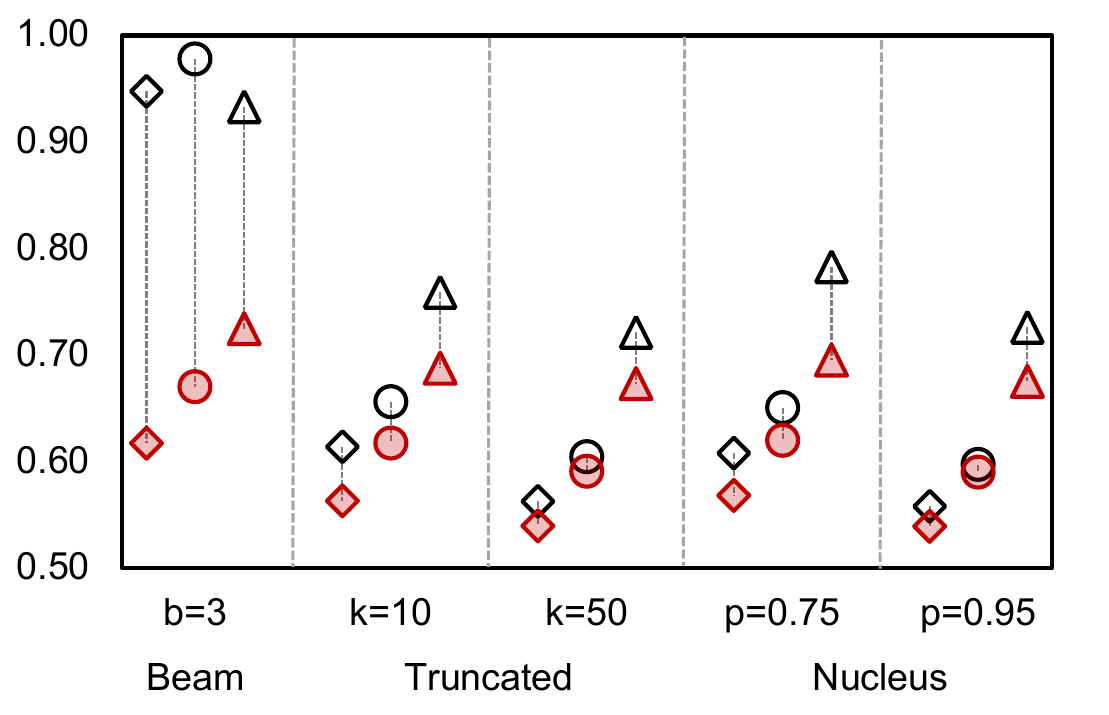}}
    \vline
    \subfigure[Entropy-4($\Uparrow$)]
    {\includegraphics[width=0.328\textwidth]{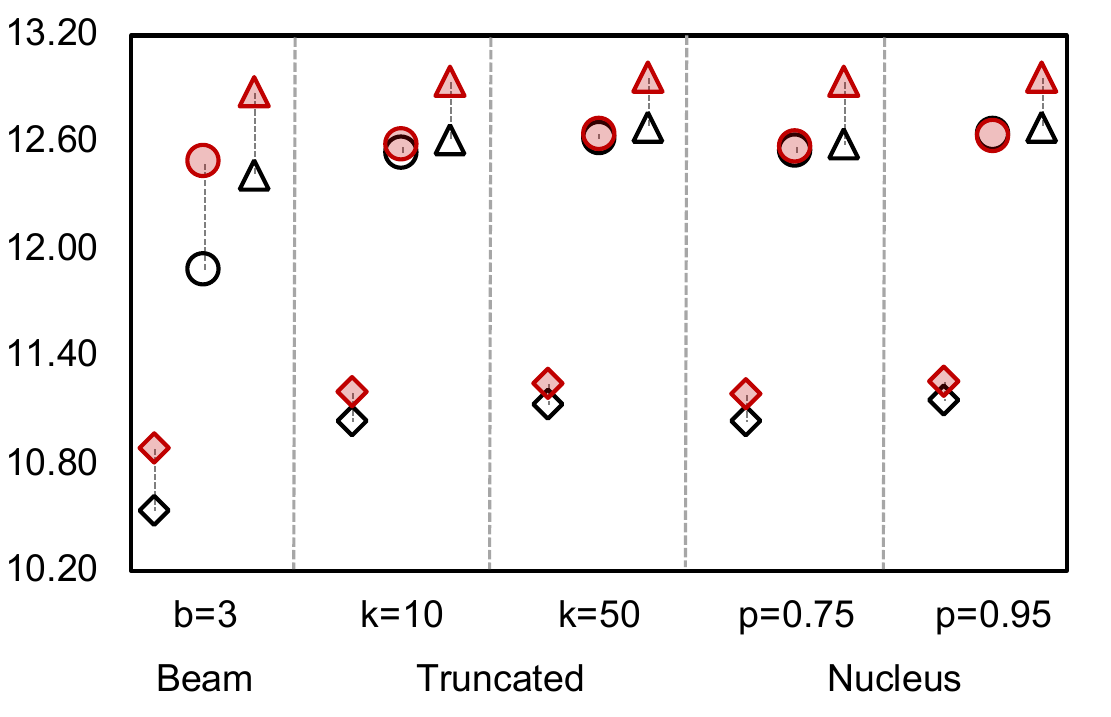}}
    \vline
    \subfigure[Oracle-BLEU-4($\Uparrow$)]
    {\includegraphics[width=0.328\textwidth]{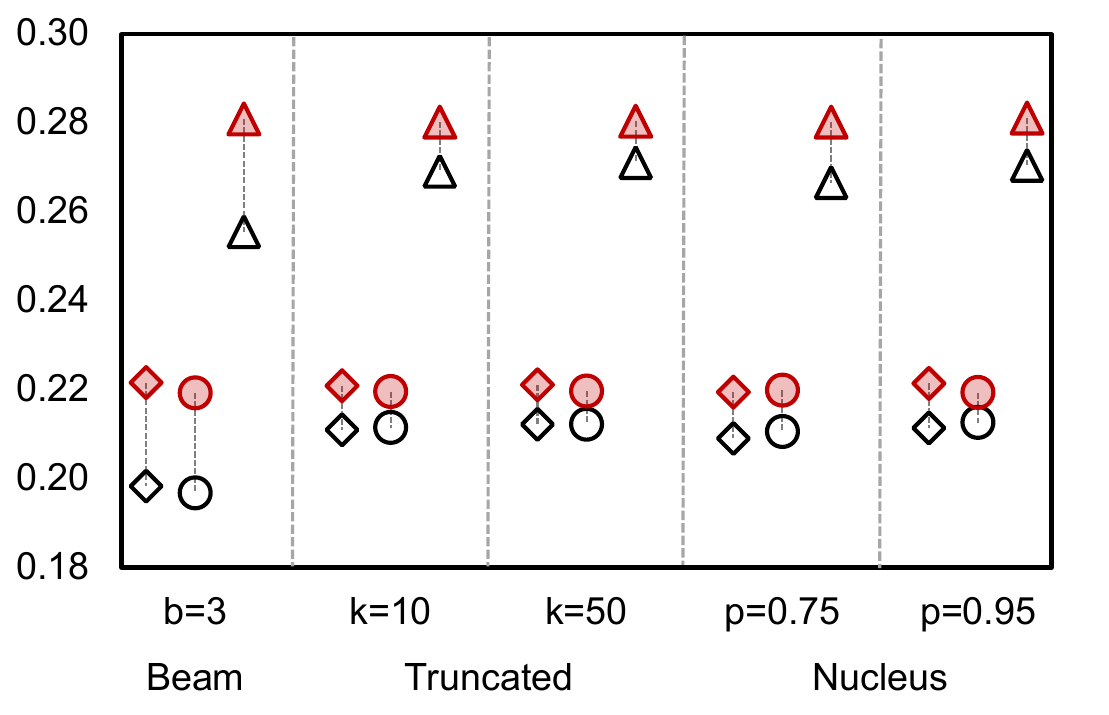}}
    \vspace{-0.15in}
    \caption{\textit{PermGen} demonstrates superior performance on both diversity and accuracy compared with different diversity-promoting methods. The specific values involved in the figure are shown in Table \ref{tab:appendix-top1} in Appendix.}
    \label{fig:diversity-and-acc}
\end{figure*}

\begin{figure*}[t]
    \centering
    {\includegraphics[width=1.0\textwidth]{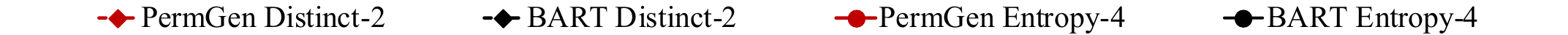}}
    \subfigure[ROCStories]
    {\includegraphics[width=0.3285\textwidth]{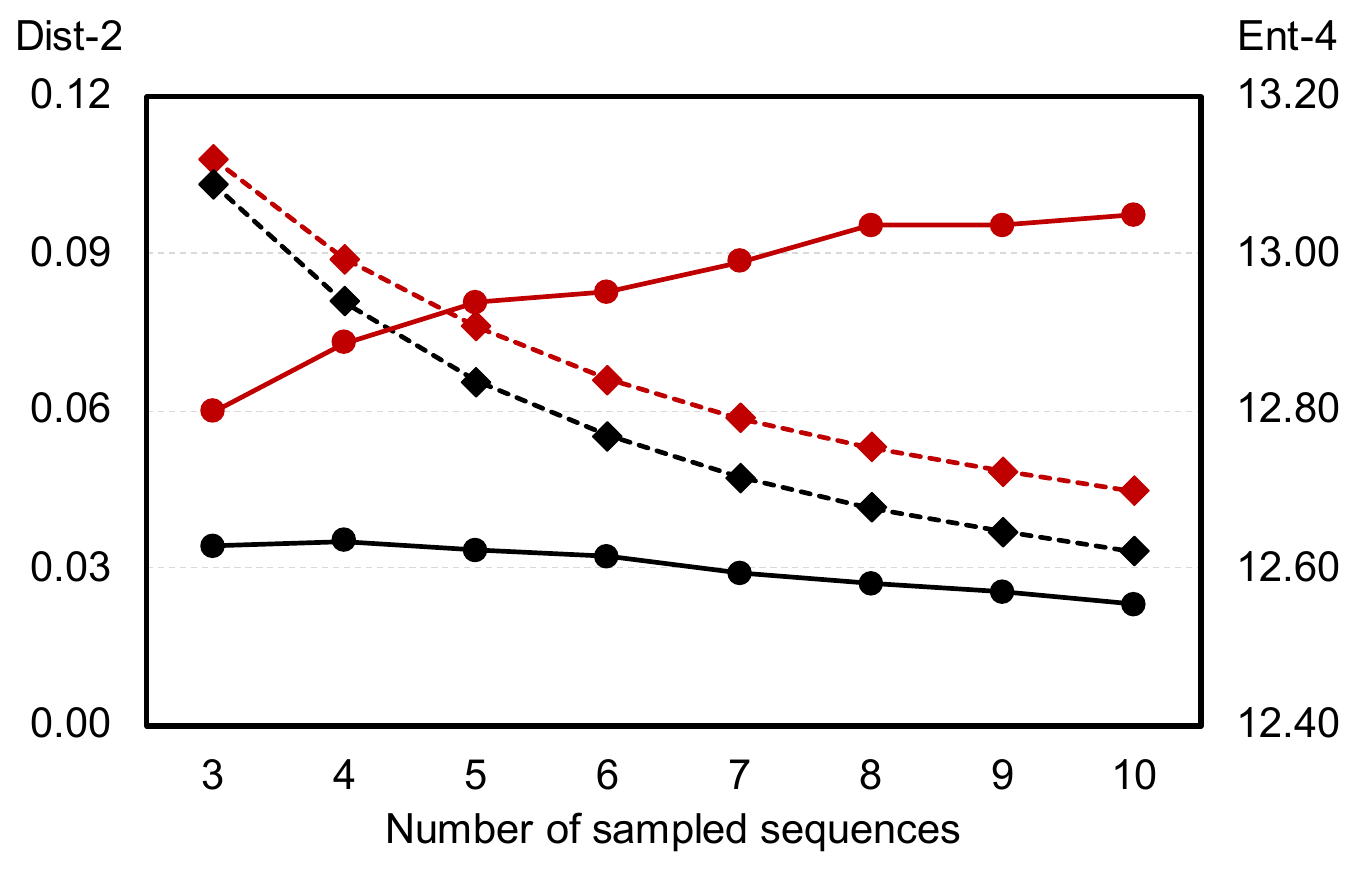}}
    \vline
    \subfigure[AGENDA]
    {\includegraphics[width=0.3285\textwidth]{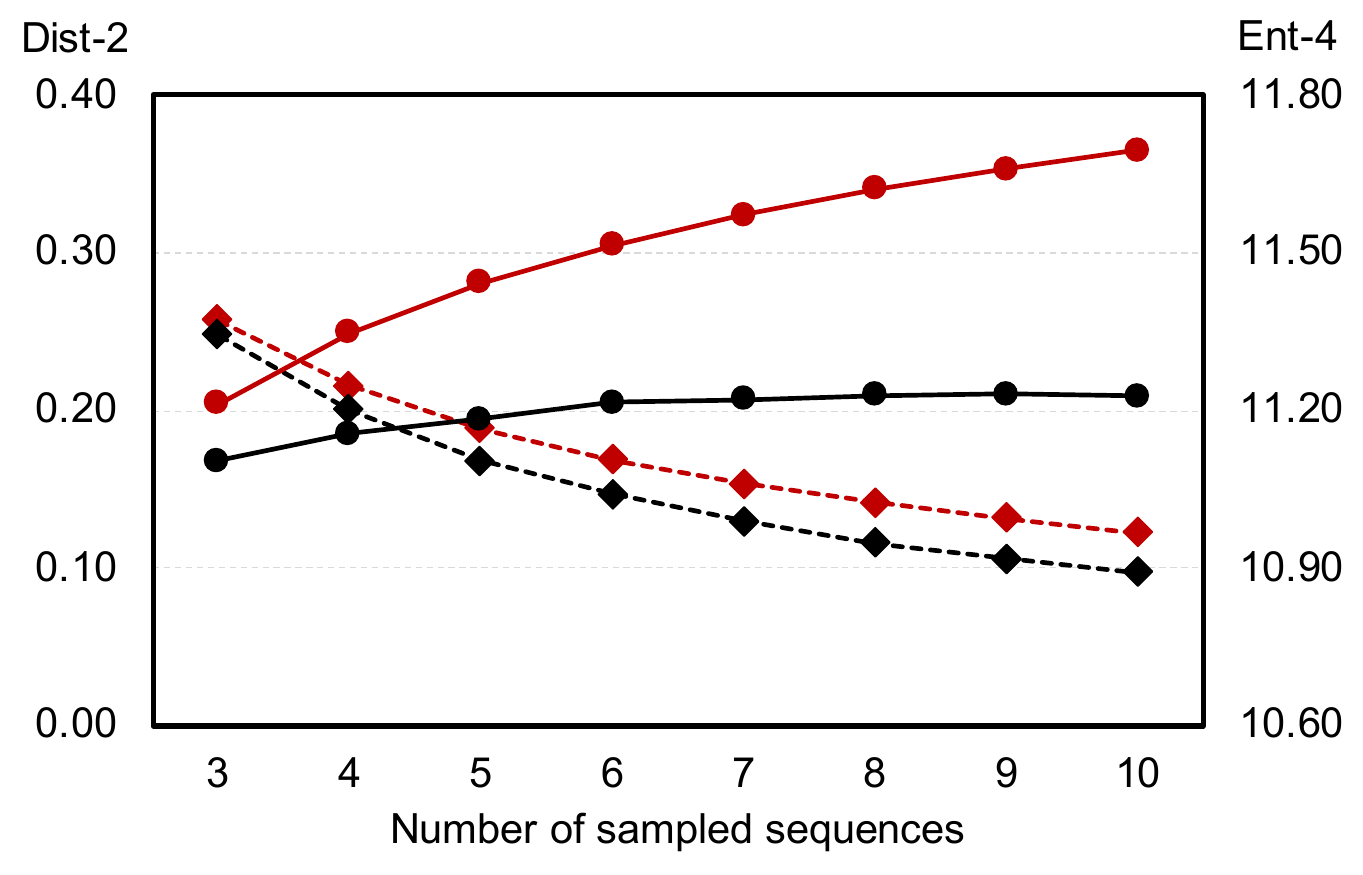}}
    \vline
    \subfigure[DailyMail]
    {\includegraphics[width=0.3285\textwidth]{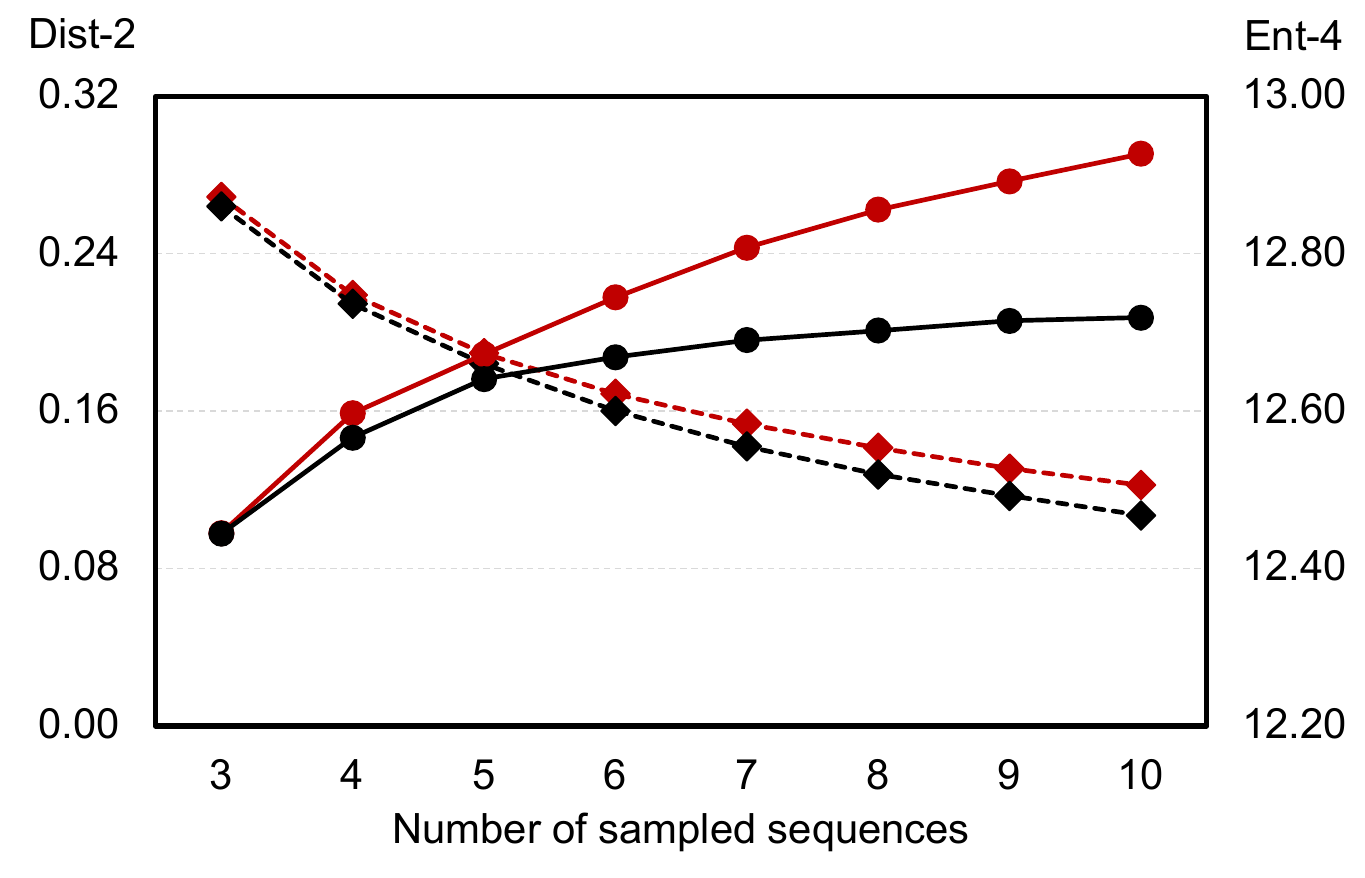}}
    \vspace{-0.15in}
    \caption{\textit{PermGen} generates more diverse paragraphs over different number of sampled candidates. The diversity measured at each point is the mean value of Dist-$2$ and Ent-$4$ when $k=10$, $k=50$, $p=.75$, and $p=.95$.}
    \label{fig:diverse}
\end{figure*}

\subsection{Baseline Methods}

We compared with three pre-trained Transformer-based models: BART~\cite{lewis2019bart}, T5~\cite{raffel2020exploring} and BERTGen~\cite{rothe2020leveraging}. These models have demonstrated state-of-the-art performance in various tasks. We also compare with GPT-2~\cite{radford2019language}  
and two recent non-autoregressive generation models: BLM~\cite{shen2020blank} and POINTER~\cite{zhang2020pointer}.

\paragraph{BLM}~\cite{shen2020blank} Blank Language Model (BLM) generates sequences by dynamically creating and filling in blanks. The blanks control which part of the sequence to fill out, making it ideal for word-to-sequence expansion tasks.

\paragraph{POINTER}~\cite{zhang2020pointer} POINTER operates by progressively inserting new tokens between existing tokens in a \textit{parallel} manner. This procedure is recursively applied until a sequence is completed. This coarse-to-fine hierarchy makes the generation process intuitive and interpretable.

\paragraph{}

For each task, we also evaluate \textit{PermGen} with different sampling methods for decoding, including beam search, Truncated sampling~\cite{fan2018hierarchical} and Nucleus sampling~\cite{holtzman2020curious}. 

\paragraph{Truncated Sampling}~\cite{fan2018hierarchical} It randomly samples words from top-k candidates of the distribution at the decoding step.

\paragraph{Nucleus Sampling}~\cite{holtzman2020curious}  It avoids text degeneration by truncating the unreliable tail of the probability distribution, sampling from the dynamic nucleus of tokens containing the vast majority of the probability mass.

\subsection{Implementation Details}

We use pre-trained parameters from BART-base~\cite{lewis2019bart} to initialized our model, which takes a maximum 512 input token sequence and consists of a 6-layer transformer encoders and another 6-layer transformer decoders~\cite{vaswani2017attention} with 12 attention heads and 768 word dimensions. For model fine tuning, we use Adam with learning rate of 3e-5, $\beta_1$ = 0.9, $\beta_2$ = 0.999, L2 weight decay of 0.01, learning rate warm up over the first 10,000 steps, and linear decay of learning rate. Our models are trained with a 4-card 32GB memory Tesla V100 GPU, and implemented with the Huggingface's Transformer~\cite{wolf2020transformers}. 

\vspace{-0.05in}
\subsection{Evaluation Metrics}

We use metrics introduced in previous work~\cite{ott2018analyzing,vijayakumar2018diverse,zhu2018texygen} to evaluate accuracy and diversity.

\vspace{-0.05in}
\subsubsection{Accuracy metrics}
\paragraph{Top-1 metric ($\Uparrow$).} This measures the Top-1 accuracy among the generated hypotheses. The accuracy is measured using corpus-level metrics, including BLEU~\cite{papineni2002bleu}, METEOR~\cite{banerjee2005meteor}, and CIDEr~\cite{vedantam2015cider}.

\vspace{-0.05in}
\paragraph{Oracle metric ($\Uparrow$).} This measures the highest accuracy comparing the best hypothesis among the top-$K$ with the target~\cite{vijayakumar2018diverse}. 

\vspace{-0.05in}
\subsubsection{Diversity metrics}
\paragraph{Corpus diversity ($\Uparrow$).}

Distinct-$k$~\cite{li2016diversity} measures the total number of unique $k$-grams normalized by the total number of generated $k$-gram tokens to avoid favoring long sentences. Entropy-$k$~\cite{zhang2018generating} reflects how evenly the empirical $k$-gram distribution is for a given sentence when word frequency is taken into account (i.e. low weights for high-frequency words).

\vspace{-0.05in}
\paragraph{Pairwise diversity ($\Downarrow$).} Referred as ``self-'' (e.g., self-BLEU)~\cite{zhu2018texygen}, it measures the within-distribution similarity. This metric computes the average of sentence-level metrics between all pairwise combinations of hypotheses $\{{Y^{(1)}, \cdots, Y^{(K)}}\}$ generated from each source sequence $X$. Lower pairwise metric indicates
high diversity between generated hypotheses.

\begin{table*}[t]
\caption{Case study. \textit{PermGen} produces more diverse stories than beam search and nucleus sampling. We shade parts of the generated text which are distinct from other candidates. We provide more case studies in Appendix.}
\vspace{-0.1in}
\scalebox{0.81}{
{\begin{tabular}{p{19.6cm}}
\toprule
$\bullet$ \textbf{Inputs:} (Title) Mounting popularity ; (Keywords) started, company, friends, hard, year, slogging, reward, traction, excited \\
\midrule
$\bullet$ \textbf{Beam search-1:} I started a new company with some friends . It was hard at first . After a year of slogging , I got a reward . The reward was a lot of traction . Now we are all excited \hlt{to start working together .} \\
$\bullet$ \textbf{Beam search-2:} I started a new company with some friends . It was hard at first . After a year of slogging , I got a reward . The reward was a lot of traction . Now we are all excited . \\
$\bullet$ \textbf{Beam search-3:} I started a new company with some friends . It was hard at first . After a year of slogging , I got a reward . The reward was a lot of traction . We are excited \hlt{to keep doing this .} \\
\midrule
$\bullet$ \textbf{Nucleus sampling-1:} I started a new company with some friends . It was hard at first . After a year of slogging , the reward was a lot of traction . \hlt{Now we are doing really well .} I am excited \hlt{to start working with my friends .} \\
$\bullet$ \textbf{Nucleus sampling-2:} I started a new company with some friends . It was hard at first . After a year of slogging , I got a lot of reward . The reward was a lot of traction . Now we are all excited \hlt{to start working together .} \\
$\bullet$\textbf{ Nucleus sampling-3:}  I started a new company with \hlt{my} friends . It was hard at first . After a year of slogging , we got a lot of reward . \hlt{We got traction and are doing really well .}  We are excited \hlt{to keep doing this .} \\
\midrule
$\bullet$ \textbf{PermGen-1 (reordered from [2, 1, 3, 5, 4]*):} I started a new company with some friends . \hlt{I tried really hard for almost a year .} \hlt{It took a lot of slogging , but as a reward I got traction .} \hlt{Now we are all doing it together .} I 'm excited \hlt{to be doing it again .} \\
$\bullet$ \textbf{PermGen-2 (reordered from [3, 1, 2, 5, 4]):} I started a new company with some friends . \hlt{It 's been hard .} \hlt{I 've been slogging through it as a reward for getting traction .} \hlt{My friends are really excited .} I 'm excited \hlt{to see what it 's all about .} \\ 
$\bullet$ \textbf{PermGen-3 (reordered from [5, 1, 2, 3, 4]):} I started a new company with some \hlt{of my} friends . \hlt{It was hard at first .} \hlt{After a year of slogging , I got a lot of reward .} \hlt{I have many traction on social media .} I am excited \hlt{to start working with my friends .} \\
\bottomrule
* ``Reordered from [2, 1, 3, 5, 4]'' means that PermGen first generates the 2$^{\text{nd}}$ sentence, and then generates the 1$^{\text{st}}$ sentence, and so on. Finally, we reorder the generated story according to the ascending order of sentence index as shown in Figure \ref{fig:inference}.
\end{tabular}}}
\label{tab:case1}
\end{table*}

\begin{table}[t]
\begin{center}
\caption{Human Evaluations on ROCStories: \textit{PermGen} (ours) v.s. three baseline methods based on \textit{diversity}. }
\vspace{-0.1in}
\scalebox{0.85}{\begin{tabular}{l||ccc}
\toprule
 & Win & Lose & Tie  \\
\midrule
{\multirow{1}*{\textit{PermGen}}}  & \textbf{64.00\%} & 14.00\% & 22.00\% \\
{\multirow{1}*{v.s. Beam}} & ($\pm$12.71\%) & ($\pm$7.70\%) & ($\pm$10.73\%)\\
\midrule
{\multirow{1}*{\textit{PermGen}}} & \textbf{54.80\%} & 8.80\% & 36.40\% \\
{\multirow{1}*{v.s. Truncated}} & ($\pm$4.10\%) & ($\pm$5.31\%) & ($\pm$5.43\%) \\
\midrule
{\multirow{1}*{\textit{PermGen}}} & \textbf{56.00\%} & 11.60\% & 32.40\%  \\
{\multirow{1}*{v.s. Nucleus}} & ($\pm$8.67\%) & ($\pm$4.27\%) & ($\pm$5.57\%)\\
\bottomrule
\end{tabular}}
\vspace{-0.2in}
\label{tab:human-eval}
\end{center}
\end{table}



\begin{table}[h]
\begin{center}
\caption{Human Evaluations of \textit{PermGen} and BART on ROCStories. \textbf{Decoding algorithm is beam search.} Minimum score is 1.0, and maximum score is 5.0. }
\vspace{-0.1in}
\setlength{\tabcolsep}{3.5mm}{\scalebox{0.86}{\begin{tabular}{l||ccc}
\toprule
 & Accuracy & Fluency & Coherency \\
\midrule
BART & 3.34 & 3.93 & 3.85 \\
\textit{PermGen} & \textbf{3.42} & \textbf{3.97} & \textbf{3.88} \\
\bottomrule
\end{tabular}}}
\vspace{-0.3in}
\label{tab:human-eval-2}
\end{center}
\end{table}

\subsection{Experimental results}
\subsubsection{\textit{PermGen} v.s. Transformers}


As shown in Table \ref{tab:baseline}, \textit{PermGen} can improve both the diversity and the accuracy of generated text when initialized with either non-pretrained (Transformer) or pre-trained (BART) Transformers. For example, compared with BART which has the best performance among baselines, \textit{PermGen} reduced Self-BLEU-4 by 43.2\% and improved BLEU-4 by +1.5\% on AGENDA. And we observe similar improvement on all other paragraph generation tasks. 
More evaluation results are in Table \ref{tab:table1-value} in Appendix.

POINTER achieves the lowest performance in paragraph generation tasks. This is because its insertion operation ignores dependency between generated words so it cannot well capture the inter-sentence coherence during long-text generation. 

\textbf{It should be noted that} since BART performed the best among all baseline methods, we apply \textit{PermGen} on BART in the following evaluations.



\subsubsection{Ablation Study}
\label{sec:ablation}

As we mentioned, the absolute positions in Transformer~\cite{vaswani2017attention} of tokens might not be available when we permute the sentence orders in paragraph generation.
So, we propose the hierarchical positional embedding that consists of a global position and a local position. 
In this section, we conduct ablation study to show the adding hierarchical position embedding to BART (short as Hi-BART) does not improve diversity, compared to the original BART model. Hi-BART even underperforms than original BART (see Table \ref{tab:ablation}). This is mainly because the newly added hierarchical position embeddings are randomly initialized, without any pre-training on large corpora.

\subsubsection{\textit{PermGen} v.s. Decoding Methods}
We investigate the quality of text generated by \textit{PermGen} (built on BART) when coupled with beam search, truncated sampling and nucleus sampling. Figure \ref{fig:diversity-and-acc} shows that on average, \textit{PermGen} can significantly boost diversity by 5.81\% in Self-BLEU-3 and 6.83\% in Self-BLEU-4, respectively, and improve accuracy by +1.2\% and +1.5\% in terms of Top1-BLEU-4 and Oracle-BLEU-4.

As the diversity of generated text depends on the number of produced candidates, we compare the diversity of generation between BART and \textit{PermGen} with various number of output candidates, $K$.
Figure~\ref{fig:diverse} shows that as $K$ increases, \textit{PermGen} can consistently generate more diverse content, measured by the ratio of distinct 2-grams, Distinct-2 (dashed line). 
Meanwhile, measured by Entropy-4 (solid line), the proportion of novel words in generated candidates from \textit{PermGen} is rising as $K$ increases, while BART shows a flat or even falling trend.



\subsubsection{Human Evaluations}

We sample 100 inputs from ROCStories test set and each evaluated method generates top-3 stories. 
Every story is assigned to five annotators with NLP background.
For diversity, the annotators are given two sets of top-3 stories from two methods each time and instructed to pick the set that is more diverse. The choices are ``win,'' ``lose,'' or ``tie.''
Then, the annotators give an accuracy score from 1 to 5 to measure semantic similarity between the top-1 generated story and ground truth story. Finally, the annotators need to give a fluency and coherency score from 1 to 5 for each generated story.


Table~\ref{tab:human-eval}-\ref{tab:human-eval-2} demonstrate that \textit{PermGen} outperforms beam search in both accuracy and fluency, while significantly improving generation diversity compared with other diversity-promoting methods.



\subsubsection{Case Study}

Table \ref{tab:case1} demonstrates generated stories from different diversity-promoting methods, including beam search, nucleus sampling and our \textit{PermGen}. Overall, we observe that \textit{PermGen} can generate more diverse stories than the other two methods. We notice that stories generated by beam search often differ only by punctuation and minor morphological variations, and typically only the last sentence (or last several words) is different from others. Nucleus sampling achieves better diversity than beam search, but the stories are still following similar storylines. In comparison, \textit{PermGen} can generate semantically richer and more diverse contents.

\section{Conclusions}
\label{sec:conclusions}
In this paper, we proposed a novel sentence-permuted paragraph generation model, \textit{PermGen}. 
\textit{PermGen} maximizes the expected log likelihood of output paragraph \textit{w.r.t.} all possible sentence orders. Experiments on three paragraph generation tasks demonstrated that \textit{PermGen} outperformed original Transformer by generating more accurate and diverse text. The result is consistent on various Transformer models and decoding methods.

\section*{Acknowledgements}
This work is supported by National Science Foundation IIS-1849816 and CCF-1901059.

\balance
\bibliography{reference}

\begin{thebibliography}{51}
\expandafter\ifx\csname natexlab\endcsname\relax\def\natexlab#1{#1}\fi

\bibitem[{Banerjee and Lavie(2005)}]{banerjee2005meteor}
Satanjeev Banerjee and Alon Lavie. 2005.
\newblock Meteor: An automatic metric for mt evaluation with improved
  correlation with human judgments.
\newblock In \emph{Proceedings of the ACL workshop on Intrinsic and Extrinsic
  Evaluation Measures for Machine Translation}.

\bibitem[{Cho et~al.(2019)Cho, Seo, and Hajishirzi}]{cho2019mixture}
Jaemin Cho, Minjoon Seo, and Hannaneh Hajishirzi. 2019.
\newblock Mixture content selection for diverse sequence generation.
\newblock In \emph{Proceedings of the 2019 Conference on Empirical Methods in
  Natural Language Processing (EMNLP)}.

\bibitem[{Clark et~al.(2018)Clark, Ross, Tan, Ji, and
  Smith}]{clark2018creative}
Elizabeth Clark, Anne~Spencer Ross, Chenhao Tan, Yangfeng Ji, and Noah~A Smith.
  2018.
\newblock Creative writing with a machine in the loop: Case studies on slogans
  and stories.
\newblock In \emph{23rd International Conference on Intelligent User
  Interfaces}.

\bibitem[{Dong et~al.(2021)Dong, Yu, Zhu, and Jiang}]{dong2021injecting}
Xiangyu Dong, Wenhao Yu, Chenguang Zhu, and Meng Jiang. 2021.
\newblock Injecting entity types into entity-guided text generation.
\newblock In \emph{Conference on Empirical Methods in Natural Language
  Processing (EMNLP)}.

\bibitem[{Fan et~al.(2018)Fan, Lewis, and Dauphin}]{fan2018hierarchical}
Angela Fan, Mike Lewis, and Yann Dauphin. 2018.
\newblock Hierarchical neural story generation.
\newblock In \emph{Proceedings of the 56th Annual Meeting of the Association
  for Computational Linguistics (ACL)}.

\bibitem[{Gu et~al.(2018)Gu, Bradbury, Xiong, Li, and Socher}]{gu2018non}
Jiatao Gu, James Bradbury, Caiming Xiong, Victor~OK Li, and Richard Socher.
  2018.
\newblock Non-autoregressive neural machine translation.
\newblock \emph{International Conference for Learning Representation (ICLR)}.

\bibitem[{Guan et~al.(2019)Guan, Wang, and Huang}]{guan2019story}
Jian Guan, Yansen Wang, and Minlie Huang. 2019.
\newblock Story ending generation with incremental encoding and commonsense
  knowledge.
\newblock In \emph{Proceedings of the AAAI Conference on Artificial
  Intelligence}.

\bibitem[{Guo et~al.(2018)Guo, Lu, Cai, Zhang, Yu, and Wang}]{guo2018long}
Jiaxian Guo, Sidi Lu, Han Cai, Weinan Zhang, Yong Yu, and Jun Wang. 2018.
\newblock Long text generation via adversarial training with leaked
  information.
\newblock In \emph{Proceedings of the AAAI Conference on Artificial
  Intelligence (AAAI)}.

\bibitem[{Guo et~al.(2019)Guo, Tan, He, Qin, Xu, and Liu}]{guo2019non}
Junliang Guo, Xu~Tan, Di~He, Tao Qin, Linli Xu, and Tie-Yan Liu. 2019.
\newblock Non-autoregressive neural machine translation with enhanced decoder
  input.
\newblock In \emph{Proceedings of the AAAI Conference on Artificial
  Intelligence (AAAI)}.

\bibitem[{Gupta et~al.(2018)Gupta, Agarwal, Singh, and Rai}]{gupta2018deep}
Ankush Gupta, Arvind Agarwal, Prawaan Singh, and Piyush Rai. 2018.
\newblock A deep generative framework for paraphrase generation.
\newblock In \emph{Proceedings of the AAAI Conference on Artificial
  Intelligence}.

\bibitem[{Holtzman et~al.(2020)Holtzman, Buys, Du, Forbes, and
  Choi}]{holtzman2020curious}
Ari Holtzman, Jan Buys, Li~Du, Maxwell Forbes, and Yejin Choi. 2020.
\newblock The curious case of neural text degeneration.
\newblock \emph{International Conference for Learning Representation (ICLR)}.

\bibitem[{Hua and Wang(2019)}]{hua2019sentence}
Xinyu Hua and Lu~Wang. 2019.
\newblock Sentence-level content planning and style specification for neural
  text generation.
\newblock In \emph{Proceedings of the 2019 Conference on Empirical Methods in
  Natural Language Processing (EMNLP-IJCNLP)}.

\bibitem[{Ippolito et~al.(2019)Ippolito, Kriz, Sedoc, Kustikova, and
  Callison-Burch}]{ippolito2019comparison}
Daphne Ippolito, Reno Kriz, Joao Sedoc, Maria Kustikova, and Chris
  Callison-Burch. 2019.
\newblock Comparison of diverse decoding methods from conditional language
  models.
\newblock In \emph{Proceedings of the 57th Annual Meeting of the Association
  for Computational Linguistics (ACL)}.

\bibitem[{Koncel-Kedziorski et~al.(2019)Koncel-Kedziorski, Bekal, Luan, Lapata,
  and Hajishirzi}]{koncel2019text}
Rik Koncel-Kedziorski, Dhanush Bekal, Yi~Luan, Mirella Lapata, and Hannaneh
  Hajishirzi. 2019.
\newblock Text generation from knowledge graphs with graph transformers.
\newblock In \emph{Proceedings of the 2019 Conference of the North American
  Chapter of the Association for Computational Linguistics (NAACL-HLT)}.

\bibitem[{Lachaux et~al.(2020)Lachaux, Joulin, and Lample}]{lachaux2020target}
Marie-Anne Lachaux, Armand Joulin, and Guillaume Lample. 2020.
\newblock Target conditioning for one-to-many generation.
\newblock In \emph{Proceedings of the 2020 Conference on Empirical Methods in
  Natural Language Processing: Findings}.

\bibitem[{Lepp{\"a}nen et~al.(2017)Lepp{\"a}nen, Munezero, Granroth-Wilding,
  and Toivonen}]{leppanen2017data}
Leo Lepp{\"a}nen, Myriam Munezero, Mark Granroth-Wilding, and Hannu Toivonen.
  2017.
\newblock Data-driven news generation for automated journalism.
\newblock In \emph{Proceedings of the 10th International Conference on Natural
  Language Generation (COLING)}.

\bibitem[{Lewis et~al.(2020)Lewis, Liu, Goyal, Ghazvininejad, Mohamed, Levy,
  Stoyanov, and Zettlemoyer}]{lewis2019bart}
Mike Lewis, Yinhan Liu, Naman Goyal, Marjan Ghazvininejad, Abdelrahman Mohamed,
  Omer Levy, Ves Stoyanov, and Luke Zettlemoyer. 2020.
\newblock Bart: Denoising sequence-to-sequence pre-training for natural
  language generation, translation, and comprehension.
\newblock \emph{Proceedings of the 58th Annual Meeting of the Association for
  Computational Linguistics (ACL)}.

\bibitem[{Li et~al.(2016)Li, Galley, Brockett, Gao, and
  Dolan}]{li2016diversity}
Jiwei Li, Michel Galley, Chris Brockett, Jianfeng Gao, and Bill Dolan. 2016.
\newblock A diversity-promoting objective function for neural conversation
  models.
\newblock In \emph{Proceedings of the 2016 Conference of the North American
  Chapter of the Association for Computational Linguistics (NAACL-HLT)}.

\bibitem[{Mostafazadeh et~al.(2016)Mostafazadeh, Chambers, He, Parikh, Batra,
  Vanderwende, Kohli, and Allen}]{mostafazadeh2016corpus}
Nasrin Mostafazadeh, Nathanael Chambers, Xiaodong He, Devi Parikh, Dhruv Batra,
  Lucy Vanderwende, Pushmeet Kohli, and James Allen. 2016.
\newblock A corpus and cloze evaluation for deeper understanding of commonsense
  stories.
\newblock In \emph{Proceedings of the 2016 Conference of the North American
  Chapter of the Association for Computational Linguistics (NAACL)}.

\bibitem[{Murphy et~al.(2019)Murphy, Srinivasan, Rao, and
  Ribeiro}]{murphy2019janossy}
Ryan~L Murphy, Balasubramaniam Srinivasan, Vinayak Rao, and Bruno Ribeiro.
  2019.
\newblock Janossy pooling: Learning deep permutation-invariant functions for
  variable-size inputs.
\newblock \emph{International Conference for Learning Representation (ICLR)}.

\bibitem[{Ott et~al.(2018)Ott, Auli, Grangier, and Ranzato}]{ott2018analyzing}
Myle Ott, Michael Auli, David Grangier, and Marc’Aurelio Ranzato. 2018.
\newblock Analyzing uncertainty in neural machine translation.
\newblock In \emph{International Conference on Machine Learning (ICML)}.

\bibitem[{Papineni et~al.(2002)Papineni, Roukos, Ward, and
  Zhu}]{papineni2002bleu}
Kishore Papineni, Salim Roukos, Todd Ward, and Wei-Jing Zhu. 2002.
\newblock Bleu: a method for automatic evaluation of machine translation.
\newblock In \emph{Proceedings of the 40th annual meeting of the Association
  for Computational Linguistics (ACL)}.

\bibitem[{Puduppully et~al.(2019)Puduppully, Dong, and
  Lapata}]{puduppully2019data}
Ratish Puduppully, Li~Dong, and Mirella Lapata. 2019.
\newblock Data-to-text generation with content selection and planning.
\newblock In \emph{Proceedings of the AAAI Conference on Artificial
  Intelligence (AAAI)}.

\bibitem[{Qian et~al.(2019)Qian, Qiu, Zhang, Jiang, and Yu}]{qian2019exploring}
Lihua Qian, Lin Qiu, Weinan Zhang, Xin Jiang, and Yong Yu. 2019.
\newblock Exploring diverse expressions for paraphrase generation.
\newblock In \emph{Proceedings of the 2019 Conference on Empirical Methods in
  Natural Language Processing and the 9th International Joint Conference on
  Natural Language Processing (EMNLP-IJCNLP)}.

\bibitem[{Radford et~al.(2019)Radford, Wu, Child, Luan, Amodei, and
  Sutskever}]{radford2019language}
Alec Radford, Jeffrey Wu, Rewon Child, David Luan, Dario Amodei, and Ilya
  Sutskever. 2019.
\newblock Language models are unsupervised multitask learners.
\newblock \emph{OpenAI}.

\bibitem[{Raffel et~al.(2020)Raffel, Shazeer, Roberts, Lee, Narang, Matena,
  Zhou, Li, and Liu}]{raffel2020exploring}
Colin Raffel, Noam Shazeer, Adam Roberts, Katherine Lee, Sharan Narang, Michael
  Matena, Yanqi Zhou, Wei Li, and Peter~J Liu. 2020.
\newblock Exploring the limits of transfer learning with a unified text-to-text
  transformer.
\newblock \emph{Journal of Machine Learning Research}.

\bibitem[{Ren et~al.(2020)Ren, Liu, Tan, Zhao, Zhao, and Liu}]{ren2020study}
Yi~Ren, Jinglin Liu, Xu~Tan, Sheng Zhao, Zhou Zhao, and Tie-Yan Liu. 2020.
\newblock A study of non-autoregressive model for sequence generation.
\newblock \emph{Proceedings of the 58th annual meeting of the Association for
  Computational Linguistics (ACL)}.

\bibitem[{Robbins and Monro(1951)}]{robbins1951stochastic}
Herbert Robbins and Sutton Monro. 1951.
\newblock A stochastic approximation method.
\newblock \emph{Mathematical Statistics}.

\bibitem[{Rose et~al.(2010)Rose, Engel, Cramer, and Cowley}]{rose2010automatic}
Stuart Rose, Dave Engel, Nick Cramer, and Wendy Cowley. 2010.
\newblock Automatic keyword extraction from individual documents.
\newblock \emph{Text mining: applications and theory}.

\bibitem[{Rothe et~al.(2020)Rothe, Narayan, and Severyn}]{rothe2020leveraging}
Sascha Rothe, Shashi Narayan, and Aliaksei Severyn. 2020.
\newblock Leveraging pre-trained checkpoints for sequence generation tasks.
\newblock \emph{Transactions of the Association for Computational Linguistics
  (TACL)}.

\bibitem[{See et~al.(2017)See, Liu, and Manning}]{see2017get}
Abigail See, Peter~J Liu, and Christopher~D Manning. 2017.
\newblock Get to the point: Summarization with pointer-generator networks.
\newblock In \emph{Proceedings of the 55th Annual Meeting of the Association
  for Computational Linguistics (ACL)}.

\bibitem[{Shen et~al.(2019)Shen, Ott, Auli, and Ranzato}]{shen2019mixture}
Tianxiao Shen, Myle Ott, Michael Auli, and Marc’Aurelio Ranzato. 2019.
\newblock Mixture models for diverse machine translation: Tricks of the trade.
\newblock In \emph{International Conference on Machine Learning (ICML)}.

\bibitem[{Shen et~al.(2020)Shen, Quach, Barzilay, and Jaakkola}]{shen2020blank}
Tianxiao Shen, Victor Quach, Regina Barzilay, and Tommi Jaakkola. 2020.
\newblock Blank language models.
\newblock \emph{Proceedings of the 2020 conference on empirical methods in
  natural language processing (EMNLP)}.

\bibitem[{Shi et~al.(2018)Shi, Chen, Qiu, and Huang}]{shi2018toward}
Zhan Shi, Xinchi Chen, Xipeng Qiu, and Xuanjing Huang. 2018.
\newblock Toward diverse text generation with inverse reinforcement learning.
\newblock In \emph{Proceedings of the 27th International Joint Conference on
  Artificial Intelligence}.

\bibitem[{Stern et~al.(2019)Stern, Chan, Kiros, and
  Uszkoreit}]{stern2019insertion}
Mitchell Stern, William Chan, Jamie Kiros, and Jakob Uszkoreit. 2019.
\newblock Insertion transformer: Flexible sequence generation via insertion
  operations.
\newblock In \emph{International Conference on Machine Learning (ICML)}.

\bibitem[{Sutskever et~al.(2014)Sutskever, Vinyals, and
  Le}]{sutskever2014sequence}
Ilya Sutskever, Oriol Vinyals, and Quoc~V Le. 2014.
\newblock Sequence to sequence learning with neural networks.
\newblock In \emph{Advances in Neural Information Processing Systems
  (NeurIPS)}.

\bibitem[{Vaswani et~al.(2017)Vaswani, Shazeer, Parmar, Uszkoreit, Jones,
  Gomez, Kaiser, and Polosukhin}]{vaswani2017attention}
Ashish Vaswani, Noam Shazeer, Niki Parmar, Jakob Uszkoreit, Llion Jones,
  Aidan~N Gomez, {\L}ukasz Kaiser, and Illia Polosukhin. 2017.
\newblock Attention is all you need.
\newblock In \emph{Advances in neural information processing systems
  (NeurIPS)}.

\bibitem[{Vedantam et~al.(2015)Vedantam, Lawrence~Zitnick, and
  Parikh}]{vedantam2015cider}
Ramakrishna Vedantam, C~Lawrence~Zitnick, and Devi Parikh. 2015.
\newblock Cider: Consensus-based image description evaluation.
\newblock In \emph{Proceedings of conference on computer vision and pattern
  recognition (CVPR)}.

\bibitem[{Vijayakumar et~al.(2018)Vijayakumar, Cogswell, Selvaraju, Sun, Lee,
  Crandall, and Batra}]{vijayakumar2018diverse}
Ashwin~K Vijayakumar, Michael Cogswell, Ramprasaath~R Selvaraju, Qing Sun,
  Stefan Lee, David Crandall, and Dhruv Batra. 2018.
\newblock Diverse beam search for improved description of complex scenes.
\newblock In \emph{AAAI Conference on Artificial Intelligence}.

\bibitem[{Vijayakumar et~al.(2016)Vijayakumar, Cogswell, Selvaraju, Sun, Lee,
  Crandall, and Batra}]{vijayakumar2016diverse}
Ashwin~K Vijayakumar, Michael Cogswell, Ramprasath~R Selvaraju, Qing Sun,
  Stefan Lee, David Crandall, and Dhruv Batra. 2016.
\newblock Diverse beam search: Decoding diverse solutions from neural sequence
  models.
\newblock \emph{arXiv preprint arXiv:1610.02424}.

\bibitem[{Wang et~al.(2019)Wang, Huang, Jiang, Knight, Ji, Bansal, and
  Luan}]{wang2019paperrobot}
Qingyun Wang, Lifu Huang, Zhiying Jiang, Kevin Knight, Heng Ji, Mohit Bansal,
  and Yi~Luan. 2019.
\newblock Paperrobot: Incremental draft generation of scientific ideas.
\newblock In \emph{57th Annual Meeting of the Association for Computational
  Linguistics (ACL)}.

\bibitem[{Wang et~al.(2020)Wang, Rao, Zhang, Qin, Tian, and
  Wang}]{wang2020diversify}
Zhen Wang, Siwei Rao, Jie Zhang, Zhen Qin, Guangjian Tian, and Jun Wang. 2020.
\newblock Diversify question generation with continuous content selectors and
  question type modeling.
\newblock In \emph{Proceedings of the 2020 Conference on Empirical Methods in
  Natural Language Processing: Findings (EMNLP)}.

\bibitem[{Wolf et~al.(2020)}]{wolf2020transformers}
Thomas Wolf et~al. 2020.
\newblock Transformers: State-of-the-art natural language processing.
\newblock In \emph{Proceedings of the 2020 Conference on Empirical Methods in
  Natural Language Processing (EMNLP)}.

\bibitem[{Yang et~al.(2019)Yang, Li, Luo, Liu, and Sun}]{yang2019enhancing}
Pengcheng Yang, Lei Li, Fuli Luo, Tianyu Liu, and Xu~Sun. 2019.
\newblock Enhancing topic-to-essay generation with external commonsense
  knowledge.
\newblock In \emph{Proceedings of the 57th Annual Meeting of the Association
  for Computational Linguistics (ACL)}.

\bibitem[{Yao et~al.(2019)Yao, Peng, Weischedel, Knight, Zhao, and
  Yan}]{yao2019plan}
Lili Yao, Nanyun Peng, Ralph Weischedel, Kevin Knight, Dongyan Zhao, and Rui
  Yan. 2019.
\newblock Plan-and-write: Towards better automatic storytelling.
\newblock In \emph{Proceedings of the AAAI Conference on Artificial
  Intelligence (AAAI)}.

\bibitem[{Yu et~al.(2020)Yu, Zhu, Li, Hu, Wang, Ji, and Jiang}]{yu2020survey}
Wenhao Yu, Chenguang Zhu, Zaitang Li, Zhiting Hu, Qingyun Wang, Heng Ji, and
  Meng Jiang. 2020.
\newblock A survey of knowledge-enhanced text generation.
\newblock \emph{arXiv preprint arXiv:2010.04389}.

\bibitem[{Zhang et~al.(2018)Zhang, Galley, Gao, Gan, Li, Brockett, and
  Dolan}]{zhang2018generating}
Yizhe Zhang, Michel Galley, Jianfeng Gao, Zhe Gan, Xiujun Li, Chris Brockett,
  and Bill Dolan. 2018.
\newblock Generating informative and diverse conversational responses via
  adversarial information maximization.
\newblock \emph{Advances in Neural Information Processing Systems (NeurIPS)}.

\bibitem[{Zhang et~al.(2020)Zhang, Wang, Li, Gan, Brockett, and
  Dolan}]{zhang2020pointer}
Yizhe Zhang, Guoyin Wang, Chunyuan Li, Zhe Gan, Chris Brockett, and Bill Dolan.
  2020.
\newblock Pointer: Constrained text generation via insertion-based generative
  pre-training.
\newblock \emph{Proceedings of the 2020 conference on empirical methods in
  natural language processing (EMNLP)}.

\bibitem[{Zhao et~al.(2020)Zhao, Xu, Lin, Zhang, Yang, and Sun}]{zhao2020graph}
Liang Zhao, Jingjing Xu, Junyang Lin, Yichang Zhang, Hongxia Yang, and Xu~Sun.
  2020.
\newblock Graph-based multi-hop reasoning for long text generation.
\newblock \emph{arXiv preprint arXiv:2009.13282}.

\bibitem[{Zhao et~al.(2017)Zhao, Zhao, and Eskenazi}]{zhao2017learning}
Tiancheng Zhao, Ran Zhao, and Maxine Eskenazi. 2017.
\newblock Learning discourse-level diversity for neural dialog models using
  conditional variational autoencoders.
\newblock In \emph{Proceedings of the 55th Annual Meeting of the Association
  for Computational Linguistics}.

\bibitem[{Zhu et~al.(2018)Zhu, Lu, Zheng, Guo, Zhang, Wang, and
  Yu}]{zhu2018texygen}
Yaoming Zhu, Sidi Lu, Lei Zheng, Jiaxian Guo, Weinan Zhang, Jun Wang, and Yong
  Yu. 2018.
\newblock Texygen: A benchmarking platform for text generation models.
\newblock In \emph{The 41st International ACM SIGIR Conference on Research \&
  Development in Information Retrieval (SIGIR)}.

\end{thebibliography}
\bibliographystyle{acl_natbib}

\clearpage
\appendix

\begin{table*}[t]
\begin{center}
\caption{Extension of Table \ref{tab:baseline}. Accuracy results for \textit{PermGen} and baseline methods.}
\vspace{-0.05in}
\setlength{\tabcolsep}{3mm}{\scalebox{0.90}{\begin{tabular}{lc|ccccccc}
\toprule
{\multirow{1}*{Method}} & Pre-train & BLEU-1$\Uparrow$ & BLEU-2$\Uparrow$ & BLEU-3$\Uparrow$ & BLEU-4$\Uparrow$ &  METEOR$\Uparrow$ & CIDEr$\Uparrow$  \\
\midrule
\rowcolor{gray!12} \multicolumn{8}{c}{AGENDA} \\
\midrule
BLM & $\surd$ & 0.4632 & 0.3248 & 0.2338 & 0.1679 & 0.2282 & 0.4940 \\
GPT-2 & $\surd$ & 0.4009 & 0.2608 & 0.1795 & 0.1247 & 0.2117 & 0.6508 \\
BERTGen & $\surd$ & 0.4435 & 0.2936 & 0.2046 & 0.1462 & 0.2202 & 0.6288 \\
T5 & $\surd$ & 0.4724 & 0.3235 & 0.2314 & 0.1688 & 0.2237 & 0.6203 \\
Transformer & $\times$ & 0.4481 & 0.3084 & 0.2175 & 0.1540 & 0.2265 & 0.6307 \\
BART & $\surd$ & 0.4765 & 0.3409 & 0.2539 & 0.1922 & 0.2418 & 0.7982 \\
\midrule
{\multirow{2}*{\textit{PermGen}}} & $\surd$ & 0.4537 & 0.3183 & 0.2308 & 0.1678 & 0.2294 & 0.6192 \\
& $\surd$ & \textbf{0.5043} & \textbf{0.3627} & \textbf{0.2719} & \textbf{0.2059} & \textbf{0.2500} & \textbf{0.8175} \\
\midrule
\rowcolor{gray!12} \multicolumn{8}{c}{DailyMail} \\
\midrule
BLM & $\surd$ & 0.3902 & 0.2540 & 0.1718 & 0.1164 & 0.2013 & 0.3627 \\
GPT-2 & $\surd$ & 0.3518 & 0.2188 & 0.1510 & 0.1072 & 0.1911 & 0.5433 \\
BERTGen & $\surd$ & 0.4350 & 0.3088 & 0.2299 & 0.1728 & 0.2301 & 0.4316 \\
T5 & $\surd$ & 0.4208 & 0.2895 & 0.2089 & 0.1529 & 0.2188 & 0.3813 \\
Transformer & $\times$ & 0.4379 & 0.2958 & 0.2098 & 0.1496 & 0.2221 & 0.5869 \\
BART & $\surd$ & 0.4727 & 0.3368 & 0.2534 & 0.1935 & 0.2379 & 0.7252 \\
\midrule
{\multirow{2}*{\textit{PermGen}}} & $\times$  & 0.4484 & 0.3024 & 0.2178 & 0.1592 & 0.2225 & 0.7059 \\
& $\surd$  & \textbf{0.4764} & \textbf{0.3407} & \textbf{0.2586} & \textbf{0.1991} & \textbf{0.2390} & \textbf{0.7549} \\
\midrule
\rowcolor{gray!12} \multicolumn{8}{c}{ROCStories} \\
\midrule
BLM & $\surd$ & 0.4653 & 0.3075 & 0.2098 & 0.1477 & 0.2275 & 0.7060 \\
GPT-2 & $\surd$ & 0.3157 & 0.1824 & 0.1133 & 0.0726 & 0.1592 & 0.4458 \\
BERTGen & $\surd$ & 0.4753 & 0.3155 & 0.2184 & 0.1576 & 0.2393 & 0.8628 \\
T5 & $\surd$ & 0.5347 & 0.3655 & 0.2584 & 0.1895 & 0.2480 & 1.3108 \\
Transformer & $\times$ & 0.5180 & 0.3526 & 0.2482 & 0.1809 & 0.2457 & 1.1342 \\
BART & $\surd$ & 0.5688 & 0.4159 & 0.3143 & 0.2445 & 0.2744 & 1.6165 \\
\midrule
{\multirow{2}*{\textit{PermGen}}} & $\times$ & 0.5238 & 0.3566 & 0.2522 & 0.1848 & 0.2465 & 1.1775 \\
& $\surd$ & \textbf{0.5830} & \textbf{0.4238} & \textbf{0.3196} & \textbf{0.2482} & \textbf{0.2769} & \textbf{1.7385} \\
\bottomrule
\end{tabular}}}
\label{tab:table1-value}
\end{center}
\end{table*}

\begin{table*}[t]
\begin{center}
\caption{Extension of Table \ref{tab:baseline}. Diversity results for \textit{PermGen} and baseline methods.}
\vspace{-0.05in}
\setlength{\tabcolsep}{2.5mm}{{\scalebox{0.90}{\begin{tabular}{lc|ccccccc}
\toprule
{\multirow{1}*{Method}} & Per-train & Dist-2$\Uparrow$ & Dist-3$\Uparrow$ & Ent-1$\Uparrow$ & Ent-2$\Uparrow$ & Ent-3$\Uparrow$ & Ent-4$\Uparrow$ &  \multicolumn{1}{c}{Self-BLEU-4$\Downarrow$}\\
\midrule
\rowcolor{gray!12} \multicolumn{9}{c}{AGENDA} \\
\midrule
BLM & $\surd$ & 0.1465 & 0.2593 & 6.2633 & 9.3736 & 10.5771 & 10.8762 & 0.9396 \\
GPT-2 & $\surd$ & 0.1665 & 0.2709 & 6.3327 & 9.0065 & 10.0387 & 10.4467 & 0.9331 \\
BERTGen & $\surd$ & 0.1463 & 0.2457 & 6.1160 & 9.1015 & 10.2476 & 10.6362 & 0.9356 \\
T5 & $\surd$ & 0.1323 & 0.2178 & 5.9393 & 8.6682 & 9.8081 & 10.2888 & 0.9421 \\
Transformer & $\times$ & 0.1489 & 0.2546 & 5.9858 & 8.6806 & 9.8452 & 10.3447 & 0.9265 \\
BART & $\surd$ & 0.1697 & 0.2815 & 6.2168 & 9.0698 & 10.1473 & 10.5333 & {0.9476} \\
\midrule
{\multirow{2}*{\textit{PermGen}}} & $\times$ & 0.2203 & 0.4564 & 6.1674 & 9.0272 & 10.3163 & 10.8839 & 0.5979 \\
& $\surd$ & \textbf{0.2492} & \textbf{0.4852} & \textbf{6.3864} & \textbf{9.4862} & \textbf{10.7269} & \textbf{11.1911} & \textbf{0.6173} \\
\midrule
\rowcolor{gray!12} \multicolumn{9}{c}{DailyMail} \\
\midrule
BLM & $\surd$ & 0.0831 & 0.1379 & 6.4465 & 9.0287 & 10.3353 & 10.9684 & 0.9889 \\
GPT-2 & $\surd$ & 0.1577 & 0.2511 & 7.4221 & 10.5923 & 11.6718 & 11.9437 & 0.9287 \\
BERTGen & $\surd$ & 0.1167 & 0.1980 & 6.8704 & 10.3130 & 11.5898 & 11.8999 & 0.9744 \\
T5 & $\surd$ & 0.1086 & 0.1763 & 6.7553 & 9.7788 & 11.0747 & 11.5242 & 0.9779 \\
Transformer & $\times$ & 0.1109 & 0.1942 & 6.9249 & 10.0132 & 11.3008 & 11.7343& 0.9678 \\
BART & $\surd$ & 0.1306 & 0.2163 & 7.1025 & 10.3602 & 11.5483 & 11.8704 & {0.9778} \\
\midrule
{\multirow{2}*{\textit{PermGen}}} & $\times$ & 0.1934 & 0.3790 & 7.1776 & 10.6234 & 11.9798 & 12.3989 & 0.7757 \\
 & $\surd$ & \textbf{0.2065} & \textbf{0.4140} & \textbf{7.2142} & \textbf{10.6632} & \textbf{12.0205} & \textbf{12.4793} & \textbf{0.6701} \\
\midrule
\rowcolor{gray!12} \multicolumn{9}{c}{ROCStories} \\
\midrule
BLM & $\surd$ & 0.0560 & 0.1402 & 5.5457 & 8.8401 & 10.8404 & 11.8790 & 0.9573 \\
GPT-2 & $\surd$ & 0.0915 & 0.1902 & 6.4372 & 9.8739 & 11.4976 & 12.0918 & 0.9194 \\
BERTGen & $\surd$ & 0.0672 & 0.1626 & 5.8283 & 9.3407 & 11.3309 & 12.2154 & 0.9456\\
T5 & $\surd$ & 0.0684 & 0.1631 & 5.8398 & 9.3663 & 11.3523 & 12.2285 & 0.9403 \\
Transformer & $\times$ & 0.0806 & 0.1971 & 5.7978 & 9.3205 & 11.4086 & 12.4069 & 0.9341 \\
BART & $\surd$ & 0.0839 & 0.2103 & 5.8986 & 9.4791 & 11.5186 & 12.4204 & {0.9330} \\
\midrule
{\multirow{2}*{\textit{PermGen}}} & $\times$ & 0.0992 & 0.2786 & 5.7773 & 9.2959 & 11.4553 & 12.6124 & 0.8548 \\
 & $\surd$ & \textbf{0.1059} & \textbf{0.2953} & \textbf{5.9669} & \textbf{9.6523} & \textbf{11.7972} & \textbf{12.8034} & \textbf{0.7247} \\
\bottomrule
\end{tabular}}}}
\label{tab:table1-value-2}
\end{center}
\end{table*}

\begin{table*}[t]
\begin{center}
\caption{Extension of Figure \ref{fig:diversity-and-acc}. Diversity results of \textit{PermGen} and other sampling methods. }
\vspace{-0.1in}
\setlength{\tabcolsep}{2.6mm}{\scalebox{0.85}{\begin{tabular}{lcc|cccccccc}
\toprule
Sampling & Setting & Method & Dist-2$\Uparrow$ & Dist-3$\Uparrow$ & Ent-1$\Uparrow$ & Ent-2$\Uparrow$ & Ent-3$\Uparrow$ & Ent-4$\Uparrow$ &  \multicolumn{2}{c}{Self-BLEU-3/4$\Downarrow$}\\
\midrule
\rowcolor{gray!12} \multicolumn{11}{c}{AGENDA} \\
\midrule
{\multirow{2}*{\makecell[c]{Beam \\ search}}} & & Original & 0.1697 & 0.2815 & 6.2168 & 9.0698 & 10.1473 & 10.5333 & {0.9531} & {0.9476} \\
& & \textit{PermGen} & 0.2492 & 0.4852 & 6.3864 & 9.4862 & 10.7269 & 11.1911 & {0.6675} & {0.6173} \\
\midrule
{\multirow{4}*{\makecell[l]{Truncated \\ sampling}}} & {\multirow{2}*{k=10}} & Original & 0.2410 & 0.4489 & 6.2751 & 9.2836 & 10.5489 & 11.0561 & {0.6680} & {0.6139} \\
& & \textit{PermGen} & 0.2599 & 0.5114 & 6.3993 & 9.5307 & 10.8008 & 11.2772 & {0.6217} & {0.5630} \\
\cmidrule{2-11}
& {\multirow{2}*{k=50}} & Original & 0.2491 & 0.4923 & 6.2773 & 9.3190 & 10.6157 & 11.1347 & {0.6239} & {0.5626} \\
&  & \textit{PermGen} & 0.2632 & 0.5205 & \textbf{6.4038} & \textbf{9.5549} & \textbf{10.8411} & \textbf{11.3239} & \textbf{{0.6010}} & 0.5397  \\
\midrule
{\multirow{4}*{\makecell[l]{Nucleus \\ sampling}}} & {\multirow{2}*{p=.75}} & Original & 0.2434 & 0.4716 & 6.2771 & 9.2946 & 10.5610 & 11.0647 & {0.6619} & {0.6080} \\
& & \textit{PermGen} & 0.2583 & 0.5076 & 6.3911 & 9.5135 & 10.7804 & 11.2588 & {0.6255} & {0.5682} \\
\cmidrule{2-11}
& {\multirow{2}*{p=.95}} & Original & 0.2522 & 0.4962 & 6.2889 & 9.3377 & 10.6308 & 11.1448 & {0.6190} & {0.5579} \\
& & \textit{PermGen} & \textbf{0.2636} & \textbf{0.5221} & 6.3996 & 9.5507 & 10.8385 & 11.3204 & {0.6006} & \textbf{0.5393}  \\
\midrule
\rowcolor{gray!12} \multicolumn{11}{c}{DailyMail} \\
\midrule
{\multirow{2}*{\makecell[c]{Beam \\ search}}} &  & Original & 0.1306 & 0.2163 & 7.1025 & 10.3602 & 11.5483 & 11.8704 & {0.9798} & {0.9778} \\
& & \textit{PermGen} & 0.2065 & 0.4140 & 7.2142 & 10.6632 & 12.0205 & 12.4793 & {0.7146} & {0.6701} \\
\midrule
{\multirow{4}*{\makecell[l]{Truncated \\ sampling}}} & {\multirow{2}*{k=10}} & Original & 0.2087 & 0.4278 & 7.1406 & 10.5733 & 12.0173 & 12.5324 & {0.7055} & {0.6561} \\
& & \textit{PermGen} & 0.2166 & 0.4421 & 7.2171 & 10.6979 & 12.0915 & 12.5711 & {0.6704} & {0.6175} \\
\cmidrule{2-11}
& {\multirow{2}*{k=50}} & Original & 0.2194 & 0.4553 & 7.1485 & 10.6153 & 12.0872 & 12.6152 & {0.6612} & {0.6046} \\
&  & \textit{PermGen} & \textbf{0.2228} & \textbf{0.4573} & \textbf{7.2222} & \textbf{10.7221} & 12.1282 & 12.6154 & {0.6472} & {0.5909} \\
\midrule
{\multirow{4}*{\makecell[l]{Nucleus \\ sampling}}} & {\multirow{2}*{p=.75}} & Original & 0.2109 & 0.4315 & 7.1439 & 10.5835 & 12.0270 & 12.5400 & {0.7004} & {0.6508} \\
& & \textit{PermGen} & 0.2166 & 0.4409 & 7.2188 & 10.6994 & 12.0894 & 12.5671 & {0.6719} & {0.6198} \\
\cmidrule{2-11}
& {\multirow{2}*{p=.95}} & Original & 0.2213 & 0.4600 & 7.1523 & 10.6254 & 12.1002 & 12.6307 & {0.6544} & {0.5972} \\
& & \textit{PermGen} & 0.2222 & 0.4566 & 7.2197 & 10.7195 & \textbf{12.1300} & \textbf{12.6168} & \textbf{{0.6465}} & \textbf{{0.5902}} \\
\midrule
\rowcolor{gray!12} \multicolumn{11}{c}{ROCStories} \\
\midrule
{\multirow{2}*{\makecell[c]{Beam \\ search}}} & & Original & 0.0839 & 0.2103 & 5.8986 & 9.4791 & 11.5186 & 12.4204 & {0.9420} & {0.9330} \\
& & PermGen & 0.1059 & 0.2953 & 5.9669 & 9.6523 & 11.7972 & 12.8034 & {0.7669} & {0.7247} \\
\midrule
{\multirow{4}*{\makecell[l]{Truncated \\ sampling}}} & {\multirow{2}*{k=10}} & Original & 0.1053 & 0.2834 & 5.9041 & 9.5370 & 11.6702 & 12.6739 & {0.7955} & {0.7591} \\
& & \textit{PermGen} & 0.1099 & 0.3088 & 5.9716 & 9.6702 & 11.8335 & 12.8547 & {0.7359} & {0.6878} \\
\cmidrule{2-11}
& {\multirow{2}*{k=50}} & Original & 0.1093 & 0.2970 & 5.9094 & 9.5552 & 11.7055 & 12.7234 & {0.7633} & {0.7213} \\
&  & \textit{PermGen} & \textbf{0.1114} & \textbf{0.3141} & \textbf{5.9761} & \textbf{9.6801} & \textbf{11.8497} & \textbf{12.8732} & \textbf{{0.7235}} & \textbf{0.6734} \\
\midrule
{\multirow{4}*{\makecell[l]{Nucleus \\ sampling}}} & {\multirow{2}*{p=.75}} & Original & 0.1032 & 0.2746 & 5.9047 & 9.5309 & 11.6501 & 12.6400 & {0.8147} & {0.7829} \\
& & \textit{PermGen} & 0.1092 & 0.3069 & 5.9730 & 9.6685 & 11.8292 & 12.8459 & {0.7421} & {0.6953} \\
\cmidrule{2-11}
& {\multirow{2}*{p=.95}} & Original & 0.1088 & 0.2955 & 5.9108 & 9.5545 & 11.7011 & 12.7160 & {0.7669} & {0.7260} \\
& & \textit{PermGen} & 0.1111 & 0.3132 & 5.9747 & 9.6778 & 11.8463 & 12.8702 & 0.7252 & {0.6753} \\
\bottomrule
\end{tabular}}}
\label{tab:appendix-diversity}
\end{center}
\end{table*}

\begin{table*}[t]
\begin{center}
\caption{Extension of Figure \ref{fig:diversity-and-acc}. Top-1 accuracy results of \textit{PermGen} and other sampling methods.}
\vspace{-0.1in}
\setlength{\tabcolsep}{3mm}{\scalebox{0.85}{\begin{tabular}{lcc||cccccc}
\toprule
Sampling & Setting & Method & BLEU-1$\Uparrow$ & BLEU-2$\Uparrow$ & BLEU-3$\Uparrow$ & BLEU-4$\Uparrow$ &  METEOR$\Uparrow$ & CIDEr$\Uparrow$  \\
\midrule
\rowcolor{gray!12} \multicolumn{9}{c}{AGENDA} \\
\midrule
{\multirow{2}*{\makecell[c]{Beam \\ search}}} & & Original & 0.4765 & 0.3409 & 0.2539 & 0.1922 & 0.2418 & 0.7982 \\
& & \textit{PermGen} & 0.5043 & 0.3627 & 0.2719 & \textbf{0.2059} & 0.2500 & 0.8175 \\
\midrule
{\multirow{4}*{\makecell[l]{Truncated \\ sampling}}} & {\multirow{2}*{k=10}} & Original & 0.4718 & 0.3374 & 0.2513 & 0.1901 & 0.2408 & 0.7191 \\
& & \textit{PermGen} & 0.5035 & 0.3620 & 0.2716 & 0.2058 & 0.2495 & 0.8317 \\
\cmidrule{2-9}
& {\multirow{2}*{k=50}} & Original & 0.4777 & 0.3403 & 0.2529 & 0.1910 & 0.2415 & 0.7874 \\
&  & \textit{PermGen} & 0.5062 & 0.3626 & 0.2709 & 0.2047 & 0.2496 & \textbf{0.8465}\\
\midrule
{\multirow{4}*{\makecell[l]{Nucleus \\ sampling}}} & {\multirow{2}*{p=.75}} & Original & 0.4720 & 0.3361 & 0.2496 & 0.1878 & 0.2399 & 0.7174 \\
& & \textit{PermGen} & 0.5021 & 0.3611 & 0.2705 & 0.2047 & 0.2487 & 0.8254 \\
\cmidrule{2-9}
& {\multirow{2}*{p=.95}} & Original & 0.4764 & 0.3391 & 0.2519 & 0.1902 & 0.2407 & 0.7818 \\
& & \textit{PermGen} & \textbf{0.5071} & \textbf{ 0.3639} &\textbf{ 0.2721} & 0.2053 & \textbf{0.2502} & 0.8114 \\
\midrule
\rowcolor{gray!12} \multicolumn{9}{c}{DailyMail} \\
\midrule
{\multirow{2}*{\makecell[c]{Beam \\ search}}} & & Original & 0.4727 & 0.3368 & 0.2534 & 0.1935 & 0.2379 & 0.7252 \\
& & \textit{PermGen} & 0.4764 & 0.3407 & 0.2586 & 0.1991 & \textbf{0.2390} & 0.7549 \\
\midrule
{\multirow{4}*{\makecell[l]{Truncated \\ sampling}}} & {\multirow{2}*{k=10}} & Original & 0.4728 & 0.3357 & 0.2522 & 0.1923 & 0.2385 & 0.7206 \\
& & \textit{PermGen} & 0.4772 & 0.3404 & 0.2578 & 0.1978 & 0.2374 & 0.8080 \\
\cmidrule{2-9}
& {\multirow{2}*{k=50}} & Original & 0.4742 & 0.3362 & 0.2519 & 0.1916 & 0.2379 & 0.7311 \\
&  & \textit{PermGen} & \textbf{0.4785} & 0.3409 & 0.2580 & 0.1982 & 0.2380 & \textbf{0.8389} \\
\midrule
{\multirow{4}*{\makecell[l]{Nucleus \\ sampling}}} & {\multirow{2}*{p=.75}} & Original & 0.4726 & 0.3360 & 0.2525 & 0.1926 & 0.2383 & 0.7070 \\
& & \textit{PermGen} & 0.4784 & \textbf{0.3416} & \textbf{0.2592} & \textbf{0.1993} & 0.2385 & 0.8048 \\
\cmidrule{2-9}
& {\multirow{2}*{p=.95}} & Original & 0.4740 & 0.3367 & 0.2528 & 0.1926 & 0.2384 & 0.7204 \\
& & \textit{PermGen} & \textbf{0.4785} & 0.3412 & 0.2584 & 0.1983 & 0.2383 & 0.8118 \\
\midrule
\rowcolor{gray!12} \multicolumn{9}{c}{ROCStories} \\
\midrule
{\multirow{2}*{\makecell[c]{Beam \\ search}}} & & Original & 0.5688 & 0.4159 & 0.3143 & 0.2445 & 0.2744 & 1.6165 \\
& & \textit{PermGen}  & 0.5830 & 0.4238 & 0.3196 & 0.2482 & 0.2769 & 1.7385 \\
\midrule
{\multirow{4}*{\makecell[l]{Truncated \\ sampling}}} & {\multirow{2}*{k=10}} & Original & 0.5680 & 0.4139 & 0.3117 & 0.2418 & 0.2734 & 1.6213 \\
& & \textit{PermGen} & 0.5829 & 0.4226 & 0.3179 & 0.2463 & 0.2761 & 1.7322 \\
\cmidrule{2-9}
& {\multirow{2}*{k=50}} & Original & 0.5681 & 0.4135 & 0.3112 & 0.2410 & 0.2729 & 1.6168 \\
&  & \textit{PermGen}  & \textbf{0.5834} & \textbf{0.4231} & 0.3181 & 0.2464 & \textbf{0.2764} & \textbf{1.7379} \\
\midrule
{\multirow{4}*{\makecell[l]{Nucleus \\ sampling}}} & {\multirow{2}*{p=.75}} & Original & 0.5681 & 0.4144 & 0.3123 & 0.2423 & 0.2735 & 1.6152 \\
& & \textit{PermGen} & 0.5823 & 0.4225 & 0.3180 & \textbf{0.2466} & 0.2761 & 1.7330 \\
\cmidrule{2-9}
& {\multirow{2}*{p=.95}} & Original & 0.5678 & 0.4135 & 0.3115 & 0.2415 & 0.2730 & 1.6207 \\
& & \textit{PermGen}  & 0.5833 & \textbf{0.4231} & \textbf{0.3183} & \textbf{0.2466} & 0.2762 & 1.7363 \\
\bottomrule
\end{tabular}}}
\label{tab:appendix-top1}
\end{center}
\end{table*}

\begin{table*}[t]
\begin{center}
\caption{Extension of Figure \ref{fig:diversity-and-acc}. Oracle (top-K) accuracy results of \textit{PermGen} and other sampling methods.}
\vspace{-0.1in}
\setlength{\tabcolsep}{3mm}{\scalebox{0.85}{\begin{tabular}{lcc||cccccc}
\toprule
Sampling & Setting & Method  & BLEU-1$\Uparrow$ & BLEU-2$\Uparrow$ & BLEU-3$\Uparrow$ & BLEU-4$\Uparrow$ &  METEOR$\Uparrow$ & CIDEr$\Uparrow$  \\
\midrule
\rowcolor{gray!12} \multicolumn{9}{c}{AGENDA} \\
\midrule
{\multirow{2}*{\makecell[c]{Beam \\ search}}} & & Original & 0.4941 & 0.3529 & 0.2624 & 0.1983 & 0.2469 & 1.0208 \\
& & \textit{PermGen} & 0.5052 & 0.3723 & 0.2858 & \textbf{0.2217} & 0.2556 & 1.0371 \\
\midrule
{\multirow{4}*{\makecell[l]{Truncated \\ sampling}}} & {\multirow{2}*{k=10}} & Original & 0.4983 & 0.3627 & 0.2746 & 0.2110 & 0.2516 & 1.0291 \\
& & \textit{PermGen} & 0.5068 & 0.3723 & 0.2852 & 0.2209 & \textbf{0.2561} & 1.0247 \\
\cmidrule{2-9}
& {\multirow{2}*{k=50}} & Original & 0.5017 & 0.3642 & 0.2758 & 0.2123 & 0.2522 & 1.0451 \\
&  & \textit{PermGen} & 0.5061 & 0.3720 & 0.2851 & 0.2211 & 0.2548 & 1.0285 \\
\midrule
{\multirow{4}*{\makecell[l]{Nucleus \\ sampling}}} & {\multirow{2}*{p=.75}} & Original & 0.4960 & 0.3604 & 0.2726 & 0.2091 & 0.2510 & 0.9920 \\
& & \textit{PermGen} & 0.5006 & 0.3688 & 0.2830 & 0.2195 & 0.2540 & \textbf{1.0549} \\
\cmidrule{2-9}
& {\multirow{2}*{p=.95}} & Original & 0.4984 & 0.3624 & 0.2744 & 0.2114 & 0.2510 & 1.0441 \\
& & \textit{PermGen} & \textbf{0.5095} & \textbf{0.3741} & \textbf{0.2863} & 0.2215 & 0.2556 & 1.0121 \\
\midrule
\rowcolor{gray!12} \multicolumn{9}{c}{DailyMail} \\
\midrule
{\multirow{2}*{\makecell[c]{Beam \\ search}}} & & Original & 0.4817 & 0.3428 & 0.2578 & 0.1968 & 0.2408 & 0.9001 \\
& & \textit{PermGen} & 0.4847 & 0.3567 & 0.2777 & 0.2193 & 0.2448 & 0.8544 \\
\midrule
{\multirow{4}*{\makecell[l]{Truncated \\ sampling}}} & {\multirow{2}*{k=10}} & Original & 0.4888 & 0.3545 & 0.2716 & 0.2115 & 0.2448 & \multicolumn{1}{c}{\textbf{0.9212}} \\
& & \textit{PermGen} & 0.4853 & 0.3568 & 0.2779 & 0.2196 & 0.2447 & 0.8678 \\
\cmidrule{2-9}
& {\multirow{2}*{k=50}} & Original & 0.4891 & 0.3550 & 0.2723 & 0.2122 & 0.2444 & 0.9197 \\
&  & \textit{PermGen} & \textbf{0.4878} & \textbf{0.3580} & \textbf{0.2784} & 0.2197 & 0.2454 & 0.8902 \\
\midrule
{\multirow{4}*{\makecell[l]{Nucleus \\ sampling}}} & {\multirow{2}*{p=.75}} & Original & 0.4888 & 0.3540 & 0.2710 & 0.2106 & 0.2444 & 0.9192 \\
& & \textit{PermGen} & 0.4857 & 0.3573 & \textbf{0.2784} & \textbf{0.2199} & 0.2451 & 0.8240 \\
\cmidrule{2-9}
& {\multirow{2}*{p=.95}} & Original & 0.4896 & 0.3554 & 0.2727 & 0.2127 & 0.2446 & 0.9152 \\
& & \textit{PermGen} & 0.4871 & \textbf{0.3580} & 0.2783 & 0.2194 & \textbf{0.2455} & 0.8809 \\
\midrule
\rowcolor{gray!12} \multicolumn{9}{c}{ROCStories} \\
\midrule
{\multirow{2}*{\makecell[c]{Beam \\ search}}} & & Original & 0.5801 & 0.4282 & 0.3264 & 0.2556 & 0.2789 & 1.7947 \\
& & \textit{PermGen}  & \textbf{0.5991} & \textbf{0.4505} & 0.3513 & 0.2810 & \textbf{0.2875} & 2.0115 \\
\midrule
{\multirow{4}*{\makecell[l]{Truncated \\ sampling}}} & {\multirow{2}*{k=10}} & Original & 0.5834 & 0.4377 & 0.3390 & 0.2692 & 0.2824 & 1.8148 \\
& & \textit{PermGen} & 0.5988 & 0.4501 & 0.3508 & 0.2803 & 0.2871 & 2.0160 \\
\cmidrule{2-9}
& {\multirow{2}*{k=50}} & Original & 0.5849 & 0.4394 & 0.3410 & 0.2712 & 0.2830 & 1.8393 \\
&  & \textit{PermGen} & 0.5989 & 0.4503 & 0.3510 & 0.2805 & 0.2872 & 2.0137 \\
\midrule
{\multirow{4}*{\makecell[l]{Nucleus \\ sampling}}} & {\multirow{2}*{p=.75}} & Original & 0.5829 & 0.4362 & 0.3369 & 0.2667 & 0.2818 & 1.8135 \\
& & \textit{PermGen} & 0.5989 & 0.4500 & 0.3508 & 0.2803 & 0.2872 & 2.0107 \\
\cmidrule{2-9}
& {\multirow{2}*{p=.95}} & Original & 0.5852 & 0.4392 & 0.3405 & 0.2706 & 0.2831 & 1.8293 \\
& & \textit{PermGen} & 0.5989 & \textbf{0.4505} & \textbf{0.3515} & \textbf{0.2812} & 0.2870 & \textbf{2.0232} \\
\bottomrule
\end{tabular}}}
\label{tab:appendix-topk}
\end{center}
\end{table*}

\begin{table*}[t]
\caption{Generation Cases on AGENDA dataset from \textit{PermGen} and other diversity promoting methods.}
\vspace{-0.05in}
\scalebox{0.83}{
{\begin{tabular}{p{18.5cm}}
\toprule
$\bullet$ \textbf{Inputs:} (Title) Hierarchical Semantic Classification: Word Sense Disambiguation with World Knowledge. (Keyphrases) learning architecture; lexical semantic classification problems; task-specific training data; background data; hierarchical learning architecture; word sense disambiguation task \\
\midrule
$\bullet$ \textbf{Beam search-1:} This paper presents a learning architecture for lexical semantic classification problems that does not require task-specific training data or background data . The learning architecture is built on top of a hierarchical learning architecture that is able to integrate both tasks-specific and background data in a principled way . We evaluate the hierarchical learning on a word sense disambiguation task and show that the hierarchical learns architecture outperforms a state-of-the-art \hlt{approach}  . \\
$\bullet$ \textbf{Beam search-2:} This paper presents a learning architecture for lexical semantic classification problems that does not require task-specific training data or background data . The learning architecture is built on top of a hierarchical learning architecture that is able to integrate both tasks-specific and background data in a principled way . We evaluate the hierarchical learning on a word sense disambiguation task and show that the hierarchical learns architecture \hlt{is competitive with} state-of-the-art \hlt{methods}  . \\
$\bullet$ \textbf{Beam search-3:} This paper presents a learning architecture for lexical semantic classification problems that does not require task-specific training data or background data . The learning architecture is built on top of a hierarchical learning architecture that is able to integrate both tasks-specific and background data in a principled way . We evaluate the hierarchical learning on a word sense disambiguation task and show that the hierarchical learns architecture outperforms the state-of-the-art  . \\
\midrule
$\bullet$ \textbf{Nucleus sampling-1:} We present a learning architecture for lexical semantic classification problems that does not require any task-specific training data or background data . The learning architecture \hlt{can be trained on a variety of tasks without any knowledge of the nature of the tasks or the sources of the training data .} \hlt{We show that this can be achieved by combining task-sparse and sparse training data ,} and show that the resulting hierarchical learning architecture outperforms state-of-the-art methods on a word sense disambiguation task . \\
$\bullet$ \textbf{Nucleus sampling-2:} We present a learning architecture for lexical semantic classification problems that does not require any task-specific training data or background data . The learning architecture \hlt{is designed to learn from both the tasks-specific and background data simultaneously .} We apply this hierarchical learning architecture to a word sense disambiguation task and show that hierarchical learning architectures outperforms state-of-the-art methods on both tasks  . \\
$\bullet$ \textbf{Nucleus sampling-3:} We present a learning architecture for lexical semantic classification problems that does not require task-specific training data or background data . \hlt{Our learning architecture is able to take advantage of both tasks by exploiting the relationship between task-sensor and background data in a principled way .} We apply our hierarchical learning architecture to a word sense disambiguation task , and show that hierarchical learning architectures outperforms state-of-the-art methods  . \\
\midrule
$\bullet$ \textbf{PermGen order-1 [1, 2, 3]:} \hlt{We present} a learning architecture for lexical semantic classification problems that does not require task-specific training data or background data . \hlt{Our learning architecture learns to partition the data in different domains into different subsets , such that the distribution of a given word is consistent across different domains .} \hlt{We show that our} hierarchical learning architecture \hlt{outperforms the state-of-the-art methods on a word sense disambiguation task , and present the results of a large-scale empirical evaluation .} \\
$\bullet$ \textbf{PermGen order-2 [2, 1, 3]:} \hlt{This paper presents} a new learning architecture for lexical semantic classification problems that \hlt{does not rely on task specific training data .} \hlt{We propose a learning architecture that leverages both task-specific and background data .} We apply this hierarchical learning architecture to a word sense disambiguation task , and show that hierarchical learning architecture outperforms state-of-the-art methods  . \\
$\bullet$ \textbf{PermGen order-3 [3, 1, 2]:} \hlt{We describe} a learning architecture for lexical semantic classification problems that does not require task-specific training data or background data . \hlt{Our learning architecture is built on top of a hierarchy of networks that learn to classify words based on their senses , and a network that ties together the tasks-specific and background data to form a hierarchy .} We apply this hierarchical learning architecture to a word sense disambiguation task , and show that hierarchical learning architecture outperforms state-of-the-art methods  . \\
\bottomrule
\end{tabular}}}
\label{tab:appendix-case-1}
\end{table*}

\begin{table*}[t]
\caption{Generation Cases on DailyMail dataset from \textit{PermGen} and other diversity promoting methods.}
\vspace{-0.05in}
\scalebox{0.83}{
{\begin{tabular}{p{18.5cm}}
\toprule
$\bullet$ \textbf{Inputs:} white british londoners; left; census figures show; white flight; equivalent; white britons; white britons; minority \\
\midrule
$\bullet$ \textbf{Beam search-1:} More than 700,000 white British Londoners have left the country in the last decade . Census figures show that a `` white flight '' has taken place in the capital since the 1990s . This is equivalent to more than one in four white Britons having left the UK . White Britons are now a minority of the population in London , according to new \hlt{figures} . \\
$\bullet$ \textbf{Beam search-2:} More than 700,000 white British Londoners have left the country in the last decade . Census figures show that a `` white flight '' has taken place in the capital since the 1990s . This is equivalent to more than one in four white Britons having left the UK . White Britons are now a minority of the population in London . \\
$\bullet$ \textbf{Beam search-3:} More than 700,000 white British Londoners have left the country in the last decade . Census figures show that a `` white flight '' has taken place in the capital since the 1990s . This is equivalent to more than one in four white Britons having left the UK . White Britons are now a minority of the population in London , according to a new \hlt{study} . \\
\midrule
$\bullet$ \textbf{Nucleus sampling-1:} More than 700,000 white British Londoners have left the country \hlt{since 2001} , according to new figures . Census figures show that the `` white flight '' of incomers has \hlt{overtaken those from elsewhere in the country .} \hlt{It} is equivalent to more than one in four white Britons having left the UK . White Britons are now a minority of the population in England \hlt{and Wales .} \\
$\bullet$ \textbf{Nucleus sampling-2:} More than 100,000 white British Londoners have left the country , according to new figures . Census figures show that a `` white flight '' of incomers has taken place in the capital . This is equivalent to more than half of all white Britons \hlt{being in the UK .} White Britons are now a minority of the population in London , \hlt{new figures have revealed .} \\
$\bullet$ \textbf{Nucleus Sampling-3:} More than 50,000 white British Londoners have left the country . Census figures show that a ‘ white flight ’ of incomers has taken place in the capital . This is equivalent to more than one in four white Britons having left the UK . White Britons are now a minority of the population in London , \hlt{according to new figures .} \\
\midrule
$\bullet$ \textbf{PermGen order-1 [1, 2, 3, 4]:} \hlt{The equivalent of} more than 10,000 white British Londoners have left the country , \hlt{it has been} revealed . Census figures show that a `` white flight '' of people \hlt{moving to the capital has overtaken London} for the first time . The equivalent of \hlt{more than 10,000 white Britons has left the country for the first time in more than a decade .} White Britons have become a minority \hlt{in the capital since 1994 , but are now only a minority .}\\
$\bullet$ \textbf{PermGen order-2 [3, 2, 4, 1]:} More than 50,000 white British Londoners have left the country for the first time , \hlt{new figures have} revealed . Census figures show that a `` white flight '' of people \hlt{from outside Europe has overtaken the city 's residents} for the first time . The equivalent of \hlt{around 40 per cent of all white Britons now live in the UK .} White Britons now make up a minority \hlt{of the population in London.} \\
$\bullet$ \textbf{PermGen order-3 [4, 2, 1, 3]:} More than 50,000 white British Londoners have left the country in the last three years . Census figures show that the number of `` white flight '' migrants has \hlt{risen by more than 50 per cent in just three years .} This is equivalent to \hlt{more than one in four white Britons in the UK .} The number of white Britons in the minority \hlt{has risen by more than 50 per cent in just three years .} \\
\bottomrule
\end{tabular}}}
\label{tab:appendix-case-3}
\end{table*}


\end{document}